\documentclass[10pt,journal,compsoc]{IEEEtran}
\usepackage{amssymb}
\usepackage{array}
\usepackage{amsfonts}
\usepackage{amsmath}
\usepackage{amsthm}
\usepackage{algorithm}
\usepackage{algorithmic}
\usepackage{booktabs}
\usepackage{bm}
\usepackage{bbm}
\usepackage{cite}
\usepackage{comment}
\usepackage{enumitem}
\usepackage{graphicx}
\usepackage{multirow}
\usepackage{times}
\usepackage{url}

\DeclareMathOperator*{\argmax}{argmax}
\DeclareMathOperator*{\argmin}{argmin}

\newcommand{\tabincell}[2]{\begin{tabular}{@{}#1@{}}#2\end{tabular}}

\begin{document}
\title{Deep Learning on Graphs: A Survey}
\author{Ziwei~Zhang, Peng~Cui and Wenwu~Zhu,~\IEEEmembership{Fellow,~IEEE}
\IEEEcompsocitemizethanks{\IEEEcompsocthanksitem Z. Zhang, P. Cui, and W. Zhu are with the Department
of Computer Science and Technology, Tsinghua University, Beijing, China. \protect\\
E-mail: zw-zhang16@mails.tsinghua.edu.cn, cuip@tsinghua.edu.cn,\protect\\wwzhu@tsinghua.edu.cn. P. Cui and W. Zhu are corresponding authors.
}}
\markboth{Journal of \LaTeX\ Class Files,~Vol.~14, No.~8, August~2015}%
{Shell \MakeLowercase{\textit{et al.}}: Bare Demo of IEEEtran.cls for Computer Society Journals}
\IEEEtitleabstractindextext{
\begin{abstract}
Deep learning has been shown to be successful in a number of domains, ranging from acoustics, images, to natural language processing. However, applying deep learning to the ubiquitous graph data is non-trivial because of the unique characteristics of graphs. Recently, substantial research efforts have been devoted to applying deep learning methods to graphs, resulting in beneficial advances in graph analysis techniques. In this survey, we comprehensively review the different types of deep learning methods on graphs. We divide the existing methods into five categories based on their model architectures and training strategies: graph recurrent neural networks, graph convolutional networks, graph autoencoders, graph reinforcement learning, and graph adversarial methods. We then provide a comprehensive overview of these methods in a systematic manner mainly by following their development history. We also analyze the differences and compositions of different methods. Finally, we briefly outline the applications in which they have been used and discuss potential future research directions.
\end{abstract}
\begin{IEEEkeywords}
Graph Data, Deep Learning, Graph Neural Network, Graph Convolutional Network, Graph Autoencoder.
\end{IEEEkeywords}}

\maketitle

\section{Introduction}
Over the past decade, deep learning has become the ``crown jewel'' of artificial intelligence and machine learning~\cite{lecun2015deep}, showing superior performance in acoustics~\cite{hinton2012deep}, images~\cite{krizhevsky2012imagenet} and natural language processing~\cite{bahdanau2015neural}, etc. The expressive power of deep learning to extract complex patterns from underlying data is well recognized. On the other hand, graphs\footnote{Graphs are also called \emph{networks} such as in social networks. In this paper, we use two terms interchangeably.} are ubiquitous in the real world, representing objects and their relationships in varied domains, including social networks, e-commerce networks, biology networks, traffic networks, and so on. Graphs are also known to have complicated structures that can contain rich underlying values~\cite{barabasi2016network}. As a result, how to utilize deep learning methods to analyze graph data has attracted considerable research attention over the past few years. This problem is non-trivial because several challenges exist in applying traditional deep learning architectures to graphs:
\begin{itemize}
\item \textbf{Irregular structures of graphs}. Unlike images, audio, and text, which have a clear grid structure, graphs have irregular structures, making it hard to generalize some of the basic mathematical operations to graphs~\cite{shuman2013emerging}. For example, defining convolution and pooling operations, which are the fundamental operations in convolutional neural networks (CNNs), for graph data is not straightforward. This problem is often referred to as the geometric deep learning problem~\cite{bronstein2017geometric}.
\item \textbf{Heterogeneity and diversity of graphs}. A graph itself can be complicated, containing diverse types and properties. For example, graphs can be heterogeneous or homogenous, weighted or unweighted, and signed or unsigned. In addition, the tasks of graphs also vary widely, ranging from node-focused problems such as node classification and link prediction to graph-focused problems such as graph classification and graph generation. These diverse types, properties, and tasks require different model architectures to tackle specific problems.
\item \textbf{Large-scale graphs}. In the big-data era, real graphs can easily have millions or billions of nodes and edges; some well-known examples are social networks and e-commerce networks~\cite{zang2016beyond}. Therefore, how to design scalable models, preferably models that have a linear time complexity with respect to the graph size, is a key problem.
\item \textbf{Incorporating interdisciplinary knowledge}. Graphs are often connected to other disciplines, such as biology, chemistry, and social sciences. This interdisciplinary nature provides both opportunities and challenges: domain knowledge can be leveraged to solve specific problems but integrating domain knowledge can complicate model designs. For example, when generating molecular graphs, the objective function and chemical constraints are often non-differentiable; therefore gradient-based training methods cannot easily be applied.
\end{itemize}

To tackle these challenges, tremendous efforts have been made in this area, resulting in a rich literature of related papers and methods. The adopted architectures and training strategies  also vary greatly, ranging from supervised to unsupervised and from convolutional to recursive. However, to the best of our knowledge, little effort has been made to systematically summarize the differences and connections between these diverse methods.

\begin{figure*}
  \centering
  \includegraphics[width = 8.3cm]{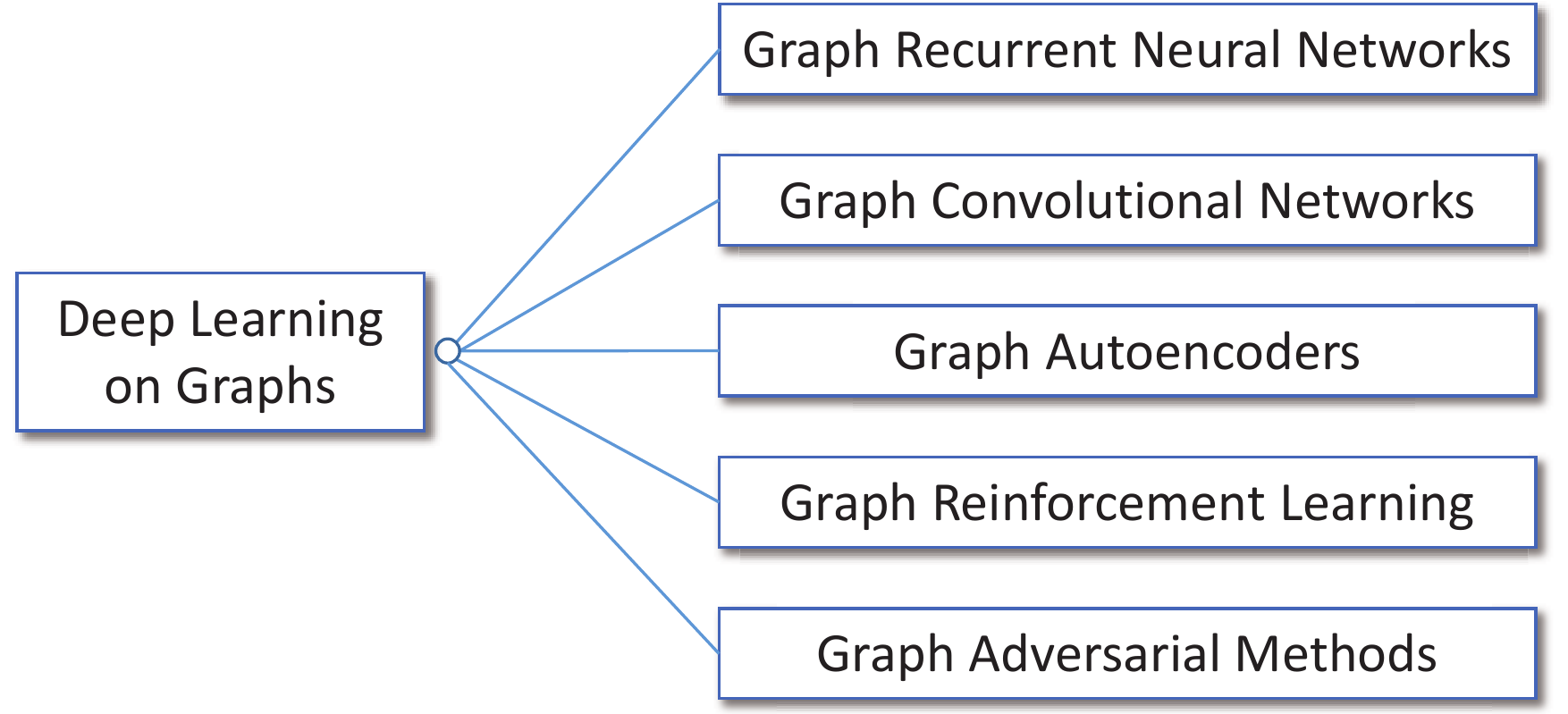}\\
  \caption{A categorization of deep learning methods on graphs. We divide the existing methods into five categories: graph recurrent neural networks, graph convolutional networks, graph autoencoders, graph reinforcement learning, and graph adversarial methods.}\label{fig:architecture}
\end{figure*}

    \begin{table*}
    \centering
    \caption{Main Distinctions among Deep Learning Methods on Graphs}\label{tab:cat}
    \begin{tabular}{ | c | c | c | c |}
    \hline
    Category                        & Basic Assumptions/Aims & Main Functions \\ \hline
    Graph recurrent neural networks & Recursive and sequential patterns of graphs & Definitions of states for nodes or graphs \\ \hline
    Graph convolutional networks    & Common local and global structural patterns of graphs & Graph convolution and readout operations \\ \hline
    Graph autoencoders              & Low-rank structures of graphs& Unsupervised node representation learning \\ \hline
    Graph reinforcement learning    & Feedbacks and constraints of graph tasks & Graph-based actions and rewards \\ \hline
    Graph adversarial methods       & The generalization ability and robustness of graph-based models & Graph adversarial trainings and attacks \\ \hline
    \end{tabular}
    \end{table*}

In this paper, we try to fill this knowledge gap by comprehensively reviewing deep learning methods on graphs. Specifically, as shown in Figure~\ref{fig:architecture}, we divide the existing methods into five categories based on their model architectures and training strategies: graph recurrent neural networks (Graph RNNs), graph convolutional networks (GCNs), graph autoencoders (GAEs), graph reinforcement learning (Graph RL), and graph adversarial methods. We summarize some of the main characteristics of these categories in Table~\ref{tab:cat}
based on the following high-level distinctions. Graph RNNs capture recursive and sequential patterns of graphs by modeling states at either the node-level or the graph-level. GCNs define convolution and readout operations on irregular graph structures to capture common local and global structural patterns. GAEs assume low-rank graph structures and adopt unsupervised methods for node representation learning. Graph RL defines graph-based actions and rewards to obtain feedbacks on graph tasks while following constraints. Graph adversarial methods adopt adversarial training techniques to enhance the generalization ability of graph-based models and test their robustness by adversarial attacks.

In the following sections, we provide a comprehensive and detailed overview of these methods, mainly by following their development history and the various ways these methods solve the challenges posed by graphs. We also analyze the differences between these models and delve into how to composite different architectures. Finally, we briefly outline the applications of these models, introduce several open libraries, and discuss potential future research directions. In the appendix, we provide a source code repository, analyze the time complexity of various methods discussed in the paper, and summarize some common applications.

\textbf{Related works}. Several previous surveys are related to our paper. Bronstein~\textit{et~al.}~\cite{bronstein2017geometric} summarized some early GCN methods as well as CNNs on manifolds and studied them comprehensively through geometric deep learning. Battaglia~\textit{et~al.}~\cite{battaglia2018relational} summarized how to use GNNs and GCNs for relational reasoning using a unified framework called graph networks, Lee~\textit{et~al.}~\cite{lee2018attention} reviewed the attention models for graphs, Zhang~\textit{et~al.}~\cite{zhang2018graph} summarized some GCNs, and Sun~\textit{et~al.}~\cite{sun2018adversarial} briefly surveyed adversarial attacks on graphs. Our work differs from these previous works in that we systematically and comprehensively review different deep learning architectures on graphs rather than focusing on one specific branch. Concurrent to our work, Zhou~\textit{et~al.}~\cite{zhou2018graph} and Wu ~\textit{et~al.}~\cite{wu2019comprehensive} surveyed this field from different viewpoints and categorizations. Specifically, neither of their works consider graph reinforcement learning or graph adversarial methods, which are covered in this paper.

Another closely related topic is network embedding, aiming to embed nodes into a low-dimensional vector space~\cite{yan2007graph,hamilton2017representation,cui2018survey}.
The main distinction between network embedding and our paper is that we focus on how different deep learning models are applied to graphs, and network embedding can be recognized as a concrete application example that uses some of these models (and it uses non-deep-learning methods as well).

The rest of this paper is organized as follows. In Section 2, we introduce the notations used in this paper and provide preliminaries. Then, we review Graph RNNs, GCNs, GAEs, Graph RL, and graph adversarial methods in Section 3 to 7, respectively. We conclude with a discussion in Section 8.

\section{Notations and Preliminaries}
\textbf{Notations}. In this paper, a graph\footnote{We consider only graphs without self-loops or multiple edges.} is represented as $G=\left(V,E \right)$ where $V=\left\{v_1,...,v_N \right\}$ is a set of $N = \left|V\right|$ nodes and $E\subseteq V \times V$ is a set of $M=\left|E \right|$ edges between nodes. We use $\mathbf{A} \in \mathbb{R}^{N\times N}$ to denote the adjacency matrix, whose $i^{th}$ row, $j^{th}$ column, and an element are denoted as $\mathbf{A}(i,:),\mathbf{A}(:,j),\mathbf{A}(i,j)$, respectively. The graph can be either directed or undirected and weighted or unweighted. In this paper, we mainly consider unsigned graphs; therefore, $\mathbf{A}(i,j) \geq 0$. Signed graphs will be discussed in future research directions. We use $\mathbf{F}^V$ and $\mathbf{F}^E$ to denote features of nodes and edges, respectively. For other variables, we use bold uppercase characters to denote matrices and bold lowercase characters to denote vectors, e.g., a matrix $\mathbf{X}$ and a vector $\mathbf{x}$. The transpose of a matrix is denoted as $\mathbf{X}^T$ and the element-wise multiplication is denoted as $\mathbf{X}_1 \odot \mathbf{X}_2$. Functions are marked with curlicues, e.g., $\mathcal{F}(\cdot)$.

To better illustrate the notations, we take social networks as an example. Each node $v_i \in V$ corresponds to a user, and the edges $E$ correspond to relations between users. The profiles of users (e.g., age, gender, and location) can be represented as node features $\mathbf{F}^V$ and interaction data (e.g., sending messages and comments) can be represented as edge features $\mathbf{F}^E$.

\begin{table}
\centering
\caption{A Table for Commonly Used Notations}\label{tab:notation}
\begin{tabular}{ | c | c | }
\hline
$G = (V,E)$                                & A graph           \\
$N,M$                                      & The number of nodes and edges \\
$V=\left\{v_1,...,v_N \right\}$            & The set of nodes         \\
$\mathbf{F}^V, \mathbf{F}^E$               & The attributes/features of nodes and edges \\
$\mathbf{A}$                               & The adjacency matrix  \\
$\mathbf{D}(i,i) = \sum_{j}\mathbf{A}(i,j)$ & The diagonal degree matrix \\
$\mathbf{L} = \mathbf{D} - \mathbf{A} $    & The Laplacian matrix  \\
$\mathbf{Q} \mathbf{\Lambda} \mathbf{Q}^T = \mathbf{L}$ & The eigendecomposition of $\mathbf{L}$ \\
$\mathbf{P} = \mathbf{D}^{-1}\mathbf{A}$   & The transition matrix   \\
$\mathcal{N}_k(i),\mathcal{N}(i)$          & The k-step and 1-step neighbors of $v_i$  \\
$\mathbf{H}^l$                             & The hidden representation in the $l^{th}$ layer \\
$f_{l}$                                    & The dimensionality of $\mathbf{H}^l$ \\
$\rho(\cdot)$                              & Some non-linear activation function\\
$\mathbf{X}_1 \odot \mathbf{X}_2$          & The element-wise multiplication \\
$\mathbf{\Theta}$                          & Learnable parameters \\
$s$                                        & The sample size \\
\hline
\end{tabular}
\end{table}

\textbf{Preliminaries}. The Laplacian matrix of an undirected graph is defined as $\mathbf{L} = \mathbf{D} - \mathbf{A}$, where $\mathbf{D} \in \mathbb{R}^{N \times N}$ is a diagonal degree matrix with $\mathbf{D}(i,i) = \sum_{j}\mathbf{A}(i,j)$. Its eigendecomposition is denoted as $\mathbf{L} = \mathbf{Q\Lambda Q^T}$, where $\mathbf{\Lambda} \in \mathbb{R}^{N \times N}$ is a diagonal matrix of eigenvalues sorted in ascending order and $\mathbf{Q} \in \mathbb{R}^{N \times N}$ are the corresponding eigenvectors. The transition matrix is defined as $\mathbf{P} = \mathbf{D}^{-1}\mathbf{A}$, where $\mathbf{P}(i,j)$ represents the probability of a random walk starting from node $v_i$ landing at node $v_j$. The $k$-step neighbors of node $v_i$ are defined as $\mathcal{N}_k(i) = \left\{j| \mathcal{D} (i,j) \leq k\right\}$, where $\mathcal{D}(i,j)$ is the shortest distance from node $v_i$ to $v_j$, i.e. $\mathcal{N}_k(i)$ is a set of nodes reachable from node $v_i$ within $k$-steps. To simplify the notation, we omit the subscript for the immediate neighborhood, i.e., $\mathcal{N}(i) = \mathcal{N}_1(i)$.

For a deep learning model, we use superscripts to denote layers, e.g., $\mathbf{H}^l$. We use $f_l$ to denote the dimensionality of the layer $l$ (i.e., $\mathbf{H}^l \in \mathbb{R}^{N\times f_l}$). The sigmoid activation function is defined as $\sigma(x) = 1 / \left( 1 + e^{-x}\right)$ and the rectified linear unit (ReLU) is defined as $\text{ReLU}(x) = max(0,x)$. A general element-wise nonlinear activation function is denoted as $\rho(\cdot)$. In this paper, unless stated otherwise, we assume all functions are differentiable, allowing the model parameters $\mathbf{\Theta}$ to be learned through back-propagation~\cite{rumelhart1986learning} using commonly adopted optimizers such as Adam~\cite{kingma2014adam} and training techniques such as dropout~\cite{Srivastava2014Dropout}. We denote the sample size as $s$ if a sampling technique is adopted. We summarize the notations in Table~\ref{tab:notation}.

The tasks for learning a deep model on graphs can be broadly divided into two categories:
\begin{itemize}
\item \textbf{Node-focused tasks}: These tasks are associated with individual nodes in the graph. Examples include node classification, link prediction, and node recommendation.
\item \textbf{Graph-focused tasks}: These tasks are associated with the entire graph. Examples include graph classification, estimating various graph properties, and generating graphs.
\end{itemize}
Note that such distinctions are more conceptually than mathematically rigorous. Some existing tasks are associated with mesoscopic structures such as community detection~\cite{wang2017community}. In addition, node-focused problems can sometimes be studied as graph-focused problems by transforming the former into egocentric networks~\cite{leskovec2012learning}. Nevertheless, we will explain the differences in algorithm designs for these two categories when necessary.

    \begin{table*}
    \centering
    \caption{The Main Characteristics of Graph Recurrent Neural Network (Graph RNNs)}\label{tab:GraphRNN}
    \begin{tabular}{ | c | c | c | c | c |}
    \hline
    Category &   Method                                          & Recursive/sequential patterns of graphs & Time Complexity  & Other Improvements \\ \hline
    \multirow{3}{*}{Node-level}  & GNN~\cite{scarselli2009graph} & \multirow{3}{*}{A recursive definition of node states} & $O(MI_f)$  & -               \\ \cline{2-2} \cline{4-5}
                                 & GGS-NNs~\cite{li2016gated}    & & $O(MT)$ & Sequence outputs \\ \cline{2-2} \cline{4-5}
                                 & SSE~\cite{dai2018learning}    & & $O(d_{\text{avg}}S)$ & -               \\ \hline

    \multirow{4}{*}{Graph-level} & You~\textit{et~al.}~\cite{you2018graphrnn}     & Generate nodes and edges in an autoregressive manner & $O(N^2)$ & -  \\ \cline{2-5}
             &   DGNN~\cite{ma2018dynamic}             & Capture the time dynamics of the formation of nodes and edges & $O(Md_{\text{avg}})$  &  -                \\ \cline{2-5}
             &   RMGCNN~\cite{monti2017geometric2}     & Recursively reconstruct the graph & $O(M)$ or $O(MN)$ & GCN layers                  \\ \cline{2-5}
             &   Dynamic GCN~\cite{manessi2017dynamic} & Gather node representations in different time slices & $O(Mt)$ & GCN layers         \\ \hline
    \end{tabular}
    \end{table*}

\section{Graph Recurrent Neural Networks}
Recurrent neural networks (RNNs) such as gated recurrent units (GRU)~\cite{cho2014learning} or long short-term memory (LSTM)~\cite{hochreiter1997long} are de facto standards in modeling sequential data. In this section, we review Graph RNNs which can capture recursive and sequential patterns of graphs. Graph RNNs can be broadly divided into two categories: node-level RNNs and graph-level RNNs. The main distinction lies in whether the patterns lie at the node-level and are modeled by node states, or at the graph-level and are modeled by a common graph state. The main characteristics of the methods surveyed are summarized in Table~\ref{tab:GraphRNN}.

\subsection{Node-level RNNs}
Node-level RNNs for graphs, which are also referred to as graph neural networks (GNNs)\footnote{Recently, GNNs have also been used to refer to general neural networks for graph data. We follow the traditional naming convention and use GNNs to refer to this specific type of Graph RNNs.}, can be dated back to the "pre-deep-learning" era~\cite{gori2005new, scarselli2009graph}. The idea behind a GNN is simple: to encode graph structural information, each node $v_i$ is represented by a low-dimensional state vector $\mathbf{s}_i$. Motivated by recursive neural networks~\cite{frasconi1998general}, a recursive definition of states is adopted~\cite{scarselli2009graph}:
\begin{equation}\label{eq:gnn1}
        \mathbf{s}_i = \sum \nolimits_{j \in \mathcal{N}(i)}\mathcal{F}(\mathbf{s}_i,\mathbf{s}_j,\mathbf{F}^V_i,\mathbf{F}^V_j,\mathbf{F}^E_{i,j} ),
\end{equation}
where $\mathcal{F}(\cdot)$ is a parametric function to be learned. After obtaining $\mathbf{s}_i$, another function $\mathcal{O}(\cdot)$ is applied to get the final outputs:
\begin{equation}\label{eq:gnn2}
        \hat y_i = \mathcal{O}( \mathbf{s}_i,\mathbf{F}^V_i ).
\end{equation}
For graph-focused tasks, the authors of~\cite{scarselli2009graph} suggested adding a special node with unique attributes to represent the entire graph.
To learn the model parameters, the following semi-supervised\footnote{It is called semi-supervised because all the graph structures and some subset of the node or graph labels is used during training.} method is adopted: after iteratively solving Eq.~\eqref{eq:gnn1} to a stable point using the Jacobi method~\cite{powell1964efficient}, one gradient descent step is performed using the Almeida-Pineda algorithm~\cite{almeida1987learning,pineda1987generalization} to minimize a task-specific objective function, for example, the squared loss between the predicted values and the ground-truth for regression tasks; then, this process is repeated until convergence.

Using the two simple equations in Eqs.~\eqref{eq:gnn1}\eqref{eq:gnn2}, GNN plays two important roles. In retrospect, a GNN unifies some of the early methods used for processing graph data, such as recursive neural networks and Markov chains~\cite{scarselli2009graph}. Looking toward the future, the general idea underlying GNNs has profound inspirations: as will be shown later, many state-of-the-art GCNs actually have a formulation similar to Eq.~\eqref{eq:gnn1} and follow the same framework of exchanging information within the immediate node neighborhoods. In fact, GNNs and GCNs can be unified into some common frameworks, and a GNN is equivalent to a GCN that uses identical layers to reach stable states. More discussion will be provided in Section 4.

Although they are conceptually important, GNNs have several drawbacks. First, to ensure that Eq.~\eqref{eq:gnn1} has a unique solution, $\mathcal{F}(\cdot)$ must be a ``contraction map''~\cite{khamsi2011introduction}, i.e., $\exists \mu,0<\mu<1$ so that
\begin{equation}
    \left\| \mathcal{F}(x) - \mathcal{F}(y) \right\| \leq \mu \left\|x - y\right\|, \forall x,y.
\end{equation}
Intuitively, a ``contraction map'' requires that the distance between any two points can only ``contract'' after the $\mathcal{F}(\cdot)$ operation, which severely limits the modeling ability. Second, because many iterations are needed to reach a stable state between gradient descend steps, GNNs are computationally expensive. Because of these drawbacks and perhaps a lack of computational power (e.g., the graphics processing unit, GPU, was not widely used for deep learning in those days) and lack of research interests, GNNs did not become a focus of general research.

A notable improvement to GNNs is gated graph sequence neural networks (GGS-NNs)~\cite{li2016gated} with the following modifications. Most importantly, the authors replaced the recursive definition in Eq.~\eqref{eq:gnn1} with a GRU, thus removing the ``contraction map'' requirement and supporting modern optimization techniques. Specifically, Eq.~\eqref{eq:gnn1} is adapted as follows:
\begin{equation}\label{eq:gnn3}
  \mathbf{s}_i^{(t)} = (1- \mathbf{z}_i^{(t)}) \odot \mathbf{s}_i^{(t-1)} + \mathbf{z}_i^{(t)} \odot \widetilde{ \mathbf{s}}_i^{(t)},
\end{equation}
where $\mathbf{z}$ is calculated by the update gate, $\widetilde{\mathbf{s}}$ is the candidate for updating, and $t$ is the pseudo time. Second, the authors proposed using several such networks operating in sequence to produce sequence outputs and showed that their method could be applied to sequence-based tasks such as program verification~\cite{brockschmidt2015learning}.

SSE~\cite{dai2018learning} took a similar approach as Eq.~\eqref{eq:gnn3}. However, instead of using a GRU in the calculation, SSE adopted stochastic fixed-point gradient descent to accelerate the training process. This scheme basically alternates between calculating steady node states using local neighborhoods and optimizing the model parameters, with both calculations in stochastic mini-batches.

\subsection{Graph-level RNNs}
In this subsection, we review how to apply RNNs to capture graph-level patterns, e.g., temporal patterns of dynamic graphs or sequential patterns at different levels of graph granularities. In graph-level RNNs, instead of applying one RNN to each node to learn the node states, a single RNN is applied to the entire graph to encode the graph states.

You~\textit{et~al.}~\cite{you2018graphrnn} applied Graph RNNs to the graph generation problem. Specifically, they adopted two RNNs: one to generate new nodes and the other to generate edges for the newly added node in an autoregressive manner. They showed that such hierarchical RNN architectures learn more effectively from input graphs than do the traditional rule-based graph generative models while having a reasonable time complexity.

To capture the temporal information of dynamic graphs, dynamic graph neural network (DGNN)~\cite{ma2018dynamic} was proposed that used a time-aware LSTM~\cite{baytas2017patient} to learn node representations. When a new edge is established, DGNN used the LSTM to update the representation of the two interacting nodes as well as their immediate neighbors, i.e., considering the one-step propagation effect. The authors showed that the time-aware LSTM could model the establishing orders and time intervals of edge formations well, which in turn benefited a range of graph applications.

\begin{figure}
\centering
\includegraphics[width = 7.1cm]{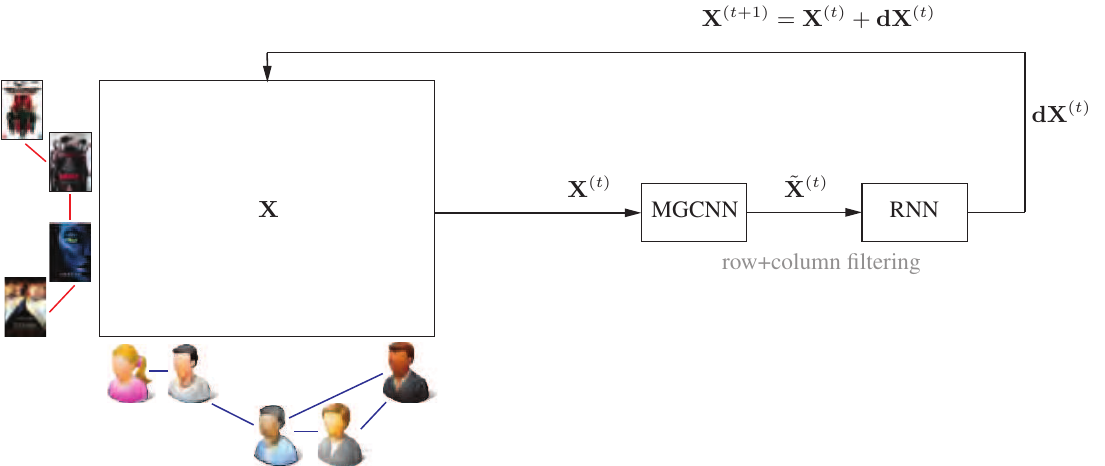}\\
\caption{The framework of RMGCNN (reprinted from~\cite{monti2017geometric2} with permission). RMGCNN includes an LSTM in the GCN to progressively reconstruct the graph. $\mathbf{X}^t$, $\mathbf{\tilde{X}}^t$, and $d\mathbf{X}^t$ represent the estimated matrix, the outputs of GCNs, and the incremental updates produced by the RNN at iteration $t$, respectively. MGCNN refers to a multigraph CNN.} \label{fig:RMGCNN}
\vspace{-0.3cm}
\end{figure}

Graph RNN can also be combined with other architectures, such as GCNs or GAEs. For example, aiming to tackle the graph sparsity problem, RMGCNN~\cite{monti2017geometric2} applied an LSTM to the results of GCNs to progressively reconstruct a graph as illustrated in Figure~\ref{fig:RMGCNN}. By using an LSTM, the information from different parts of the graph can diffuse across long ranges without requiring as many GCN layers. Dynamic GCN~\cite{manessi2017dynamic} applied an LSTM to gather the results of GCNs from different time slices in dynamic networks to capture both the spatial and temporal graph information.

    \begin{table*}
    \footnotesize
    \centering
    \caption{A Comparison among Different Graph Convolutional Networks (GCNs). T.C. = Time Complexity, M.G. = Multiple Graphs}\label{tab:GCN}
    \begin{tabular}{ | c | c | c | c | c | c | c |}
    \hline
    Method                                       & Type     & Convolution  & Readout         & T.C. & M.G. & Other Characteristics\\ \hline
    Bruna~\textit{et~al.}~\cite{bruna2014spectral}        & Spectral & Interpolation kernel & Hierarchical clustering + FC & $O(N^3)$  & No    & - \\ \hline
    Henaff~\textit{et~al.}~\cite{henaff2015deep}          & Spectral & Interpolation kernel & Hierarchical clustering + FC & $O(N^3)$  & No  & Constructing the graph \\ \hline
    ChebNet~\cite{defferrard2016convolutional}   & Spectral/Spatial  & Polynomial   & Hierarchical clustering   & $O(M)$ & Yes   & - \\ \hline
    Kipf\&Welling~\cite{kipf2017semi}            & Spectral/Spatial  & First-order  & -      & $O(M)$ & -   & - \\ \hline
    CayletNet~\cite{levie2017cayleynets}         & Spectral & Polynomial            & Hierarchical clustering + FC      & $O(M)$ & No  & - \\ \hline
    GWNN~\cite{xu2019graph}                      & Spectral & Wavelet transform     & -      & $O(M)$ & No  & - \\ \hline
    Neural FPs~\cite{duvenaud2015convolutional}  & Spatial  & First-order  & Sum             & $O(M)$ & Yes & - \\ \hline
    PATCHY-SAN~\cite{niepert2016learning}        & Spatial  & Polynomial + an order    & An order + pooling & $O(M\log N)$ & Yes   & A neighbor order \\ \hline
    LGCN~\cite{gao2018large}                     & Spatial  & First-order + an order   & -      & $O(M)$ & Yes & A neighbor order  \\ \hline
    SortPooling~\cite{zhang2018end}              & Spatial  & First-order           & An order + pooling & $O(M)$ & Yes   & A node order \\ \hline
    DCNN~\cite{atwood2016diffusion}              & Spatial  & Polynomial diffusion  & Mean   &$O(N^2)$& Yes & Edge features \\ \hline
    DGCN~\cite{zhuang2018dual}                   & Spatial  & First-order + diffusion & -    &$O(N^2)$& -   & -  \\ \hline
    MPNNs~\cite{gilmer2017neural}                & Spatial  & First-order  & Set2set         &$O(M)$  & Yes & A general framework \\ \hline
    GraphSAGE~\cite{hamilton2017inductive}       & Spatial  & First-order + sampling  & -    &$O(Ns^L)$ & Yes & A general framework \\ \hline
    MoNet~\cite{monti2017geometric}              & Spatial  & First-order  & Hierarchical clustering  & $O(M)$ & Yes   & A general framework \\ \hline
    GNs~\cite{battaglia2018relational}           & Spatial  & First-order  & A graph representation &$O(M)$& Yes   & A general framework \\ \hline
    Kearnes~\textit{et~al.}~\cite{kearnes2016molecular}   & Spatial  & Weave module & Fuzzy histogram &$O(M)$& Yes & Edge features \\ \hline
    DiffPool~\cite{ying2018hierarchical}         & Spatial  & Various  & Hierarchical clustering & $O(N^2)$ & Yes & Differentiable pooling \\ \hline
    GAT~\cite{velickovic2018graph}               & Spatial  & First-order  & -               & $O(M)$ & Yes & Attention \\ \hline
    GaAN~\cite{zhang2018gaan}                    & Spatial  & First-order  & -               & $O(Ns^L)$& Yes & Attention \\ \hline
    HAN~\cite{wang2019hetero}                    & Spatial  & Meta-path neighbors  & -       & $O(M_\phi)$ & Yes & Attention \\ \hline
    CLN~\cite{pham2017column}                    & Spatial  & First-order  & -               & $O(M)$ & -      & -  \\ \hline
    PPNP~\cite{klicpera2019predict}              & Spatial  & First-order  & -               & $O(M)$ & -     & Teleportation connections \\ \hline
    JK-Nets~\cite{xu2018representation}          & Spatial  & Various      & -               & $O(M)$ & Yes & Jumping connections \\ \hline
    ECC~\cite{simonovsky2017dynamic}             & Spatial  & First-order  & Hierarchical clustering & $O(M)$ & Yes & Edge features \\ \hline
    R-GCNs~\cite{schlichtkrull2018modeling}      & Spatial  & First-order  & -               & $O(M)$ & -   & Edge features \\ \hline
    LGNN~\cite{chen2019supervised}               & Spatial  & First-order + LINE graph &  -  & $O(M)$ & -   & Edge features \\ \hline
    PinSage~\cite{ying2018graph}                 & Spatial  & Random walk  & -               & $O(Ns^L)$& -   & Neighborhood sampling \\ \hline
    StochasticGCN~\cite{chen2018stochastic}      & Spatial  & First-order + sampling  & -    & $O(Ns^L)$& -   & Neighborhood sampling \\ \hline
    FastGCN~\cite{chen2018fastgcn}               & Spatial  & First-order + sampling  & -    & $O(NsL)$ & Yes & Layer-wise sampling \\ \hline
    Adapt~\cite{huang2018adaptive}               & Spatial  & First-order + sampling  & -    & $O(NsL)$ & Yes & Layer-wise sampling \\ \hline
    Li~\textit{et~al.}~\cite{li2018deeper}       & Spatial  & First-order  & -               & $O(M)$ & -   & Theoretical analysis \\ \hline
    SGC~\cite{wu2019simplifying}                 & Spatial  & Polynomial   & -               & $O(M)$ & Yes & Theoretical analysis \\ \hline
    GFNN~\cite{maehara2019revisiting}            & Spatial  & Polynomial   & -               & $O(M)$ & Yes & Theoretical analysis \\ \hline
    GIN~\cite{xu2019powerful}                    & Spatial  & First-order  & Sum + MLP       & $O(M)$ & Yes & Theoretical analysis \\ \hline
    DGI~\cite{velivckovic2019deep}               & Spatial  & First-order  & -               & $O(M)$ & Yes & Unsupervised training \\ \hline
    \end{tabular}
    \end{table*}
\section{Graph Convolutional Networks}
Graph convolutional networks (GCNs) are inarguably the hottest topic in graph-based deep learning. Mimicking CNNs, modern GCNs learn the common local and global structural patterns of graphs through designed convolution and readout functions. Because most GCNs can be trained with task-specific loss via backpropagation (with a few exceptions such as the unsupervised training method in~\cite{velivckovic2019deep}), we focus on the adopted architectures. We first discuss the convolution operations, then move to the readout operations and some other improvements. We summarize the main characteristics of GCNs surveyed in this paper in Table~\ref{tab:GCN}.

\subsection{Convolution Operations}\label{sec:conv}
Graph convolutions can be divided into two groups: \emph{spectral convolutions}, which perform convolution by transforming node representations into the spectral domain using the graph Fourier transform or its extensions, and \emph{spatial convolutions}, which perform convolution by considering node neighborhoods. Note that these two groups can overlap, for example, when using a polynomial spectral kernel (please refer to Section~\ref{sec:connect} for details).

\subsubsection{Spectral Methods}
Convolution is the most fundamental operation in CNNs. However, the standard convolution operation used for images or text cannot be directly applied to graphs because graphs lack a grid structure~\cite{shuman2013emerging}. Bruna~\textit{et~al.}~\cite{bruna2014spectral} first introduced convolution for graph data from the spectral domain using the graph Laplacian matrix $\mathbf{L}$~\cite{belkin2002laplacian}, which plays a similar role as the Fourier basis in signal processing~\cite{shuman2013emerging}. The graph convolution operation, $*_G$, is defined as follows:
\begin{equation}\label{eq:spectral}
    \mathbf{u}_1 *_G \mathbf{u}_2 = \mathbf{Q} \left( \left( \mathbf{Q}^T \mathbf{u}_1 \right) \odot \left( \mathbf{Q}^T \mathbf{u}_2 \right) \right),
\end{equation}
where $\mathbf{u}_1,\mathbf{u}_2 \in \mathbb{R}^N$ are two signals\footnote{We give an example of graph signals in Appendix~\ref{sec:signal}.} defined on nodes and $\mathbf{Q}$ are the eigenvectors of $\mathbf{L}$. Briefly, multiplying $\mathbf{Q}^T$ transforms the graph signals $\mathbf{u}_1,\mathbf{u}_2$ into the spectral domain (i.e., the graph Fourier transform), while multiplying $\mathbf{Q}$ performs the inverse transform. The validity of this definition is based on the convolution theorem, i.e., the Fourier transform of a convolution operation is the element-wise product of their Fourier transforms. Then, a signal $\mathbf{u}$ can be filtered by
\begin{equation}\label{eq:spectralfilter}
    \mathbf{u}' = \mathbf{Q} \mathbf{\Theta} \mathbf{Q}^T \mathbf{u},
\end{equation}
where $\mathbf{u}'$ is the output signal, $\mathbf{\Theta} = \mathbf{\Theta}(\mathbf{\Lambda})\in \mathbb{R}^{N \times N}$ is a diagonal matrix of learnable filters and $\mathbf{\Lambda}$ are the eigenvalues of $\mathbf{L}$. A convolutional layer is defined by applying different filters to different input-output signal pairs as follows:
\begin{equation}\label{eq:gcn1}
        \mathbf{u}^{l+1}_{j} = \rho \left( \sum \nolimits_{i=1}^{f_l} \mathbf{Q} \mathbf{\Theta}^l_{i,j} \mathbf{Q}^T  \mathbf{u}^l_{i}\right) \; j = 1,...,f_{l+1},
\end{equation}
where $l$ is the layer, $\mathbf{u}^l_j \in \mathbb{R}^{N}$ is the $j^{th}$ hidden representation (i.e., the signal) for the nodes in the $l^{th}$ layer, and $\mathbf{\Theta}^l_{i,j}$ are learnable filters. The idea behind Eq.~\eqref{eq:gcn1} is similar to a conventional convolution: it passes the input signals through a set of learnable filters to aggregate the information, followed by some nonlinear transformation. By using the node features $\mathbf{F}^V$ as the input layer and stacking multiple convolutional layers, the overall architecture is similar to that of a CNN. Theoretical analysis has shown that such a definition of the graph convolution operation can mimic certain geometric properties of CNNs and we refer readers to~\cite{bronstein2017geometric} for a comprehensive survey.

However, directly using Eq.~\eqref{eq:gcn1} requires learning $O(N)$ parameters, which may not be feasible in practice. Besides, the filters in the spectral domain may not be localized in the spatial domain, i.e., each node may be affected by all the other nodes rather than only the nodes in a small region. To alleviate these problems, Bruna~\textit{et~al.}~\cite{bruna2014spectral} suggested using the following smoothing filters:
\begin{equation}
    diag\left( \mathbf{\Theta}^l_{i,j} \right) = \mathcal{K} \; \alpha_{l,i,j},
\end{equation}
where $\mathcal{K}$ is a fixed interpolation kernel and $\alpha_{l,i,j}$ are learnable interpolation coefficients. The authors also generalized this idea to the setting where the graph is not given but constructed from raw features using either a supervised or an unsupervised method~\cite{henaff2015deep}.

However, two fundamental problems remain unsolved. First, because the full eigenvectors of the Laplacian matrix are needed during each calculation, the time complexity is at least $O(N^2)$ for each forward and backward pass, not to mention the $O(N^3)$ complexity required to calculate the eigendecomposition, meaning that this approach is not scalable to large-scale graphs. Second, because the filters depend on the eigenbasis $\mathbf{Q}$ of the graph, the parameters cannot be shared across multiple graphs with different sizes and structures.

Next, we review two lines of works trying to solve these limitations and then unify them using some common frameworks.

\subsubsection{The Efficiency Aspect}\label{sec:connect}
To solve the efficiency problem, ChebNet~\cite{defferrard2016convolutional} was proposed to use a polynomial filter as follows:
\begin{equation}\label{eq:cheby1}
    \mathbf{\Theta}(\mathbf{\Lambda}) = \sum \nolimits_{k=0}^{K} \theta_k \mathbf{\Lambda}^k,
\end{equation}
where $\theta_0,..., \theta_{K}$ are the learnable parameters and $K$ is the polynomial order. Then, instead of performing the eigendecomposition, the authors rewrote Eq.~\eqref{eq:cheby1} using the Chebyshev expansion~\cite{hammond2011wavelets}:
\begin{equation}\label{eq:cheby3}
    \mathbf{\Theta}(\mathbf{\Lambda}) = \sum \nolimits_{k=0}^{K} \theta_k \mathcal{T}_k ( \tilde{\mathbf{\Lambda}} ),
\end{equation}
where $\tilde{\mathbf{\Lambda}}= 2\mathbf{\Lambda} / \lambda_{max} - \mathbf{I}$ are the rescaled eigenvalues, $\lambda_{max}$ is the maximum eigenvalue, $\mathbf{I} \in \mathbb{R}^{N \times N}$ is the identity matrix, and $\mathcal{T}_k(x)$ is the Chebyshev polynomial of order $k$. The rescaling is necessary because of the orthonormal basis of Chebyshev polynomials. Using the fact that a polynomial of the Laplacian matrix acts as a polynomial of its eigenvalues, i.e., $\mathbf{L}^k = \mathbf{Q} \mathbf{\Lambda}^k \mathbf{Q}^T$, the filter operation in Eq.~\eqref{eq:spectralfilter} can be rewritten as follows:
\begin{equation}\label{eq:cheby2}
\begin{aligned}
  \mathbf{u}'  = \mathbf{Q} \mathbf{\Theta}(\mathbf{\Lambda}) \mathbf{Q}^T \mathbf{u} = & \sum \nolimits_{k=0}^{K} \theta_k \mathbf{Q} \mathcal{T}_k ( \tilde{\mathbf{\Lambda}} ) \mathbf{Q}^T \mathbf{u} \\
                                                                                      = & \sum \nolimits_{k=0}^{K} \theta_k \mathcal{T}_k ( \tilde{\mathbf{L}} ) \mathbf{u} = \sum \nolimits_{k=0}^{K} \theta_k \bar{\mathbf{u}}_k,
\end{aligned}
\end{equation}
where $\bar{\mathbf{u}}_k = \mathcal{T}_k ( \tilde{\mathbf{L}} )\mathbf{u}$ and $\tilde{\mathbf{L}} = 2\mathbf{L} / \lambda_{max} - \mathbf{I}$. Using the recurrence relation of the Chebyshev polynomial $\mathcal{T}_k(x) = 2x \mathcal{T}_{k-1}(x) - \mathcal{T}_{k-2}(x)$ and $\mathcal{T}_0(x)=1,\mathcal{T}_1(x)=x$, $\bar{\mathbf{u}}_k$ can also be calculated recursively:
\begin{equation}
    \bar{\mathbf{u}}_k = 2 \tilde{\mathbf{L}} \bar{\mathbf{u}}_{k-1} - \bar{\mathbf{u}}_{k-2}
\end{equation}
with $\bar{\mathbf{u}}_0 = \mathbf{u}$ and $\bar{\mathbf{u}}_1 = \tilde{\mathbf{L}} \mathbf{u}$. Now, because only the matrix multiplication of a sparse matrix $\tilde{\mathbf{L}}$ and some vectors need to be calculated,
the time complexity becomes $O(KM)$ when using sparse matrix multiplication, where $M$ is the number of edges and $K$ is the polynomial order, i.e., the time complexity is linear with respect to the number of edges. It is also easy to see that such a polynomial filter is strictly $K$-localized: after one convolution, the representation of node $v_i$ will be affected only by its $K$-step neighborhoods $\mathcal{N}_K(i)$. Interestingly, this idea is used independently in network embedding to preserve the high-order proximity~\cite{zhang2018arbitrary}, of which we omit the details for brevity.

\begin{figure}
\centering
\includegraphics[width = 8.5cm]{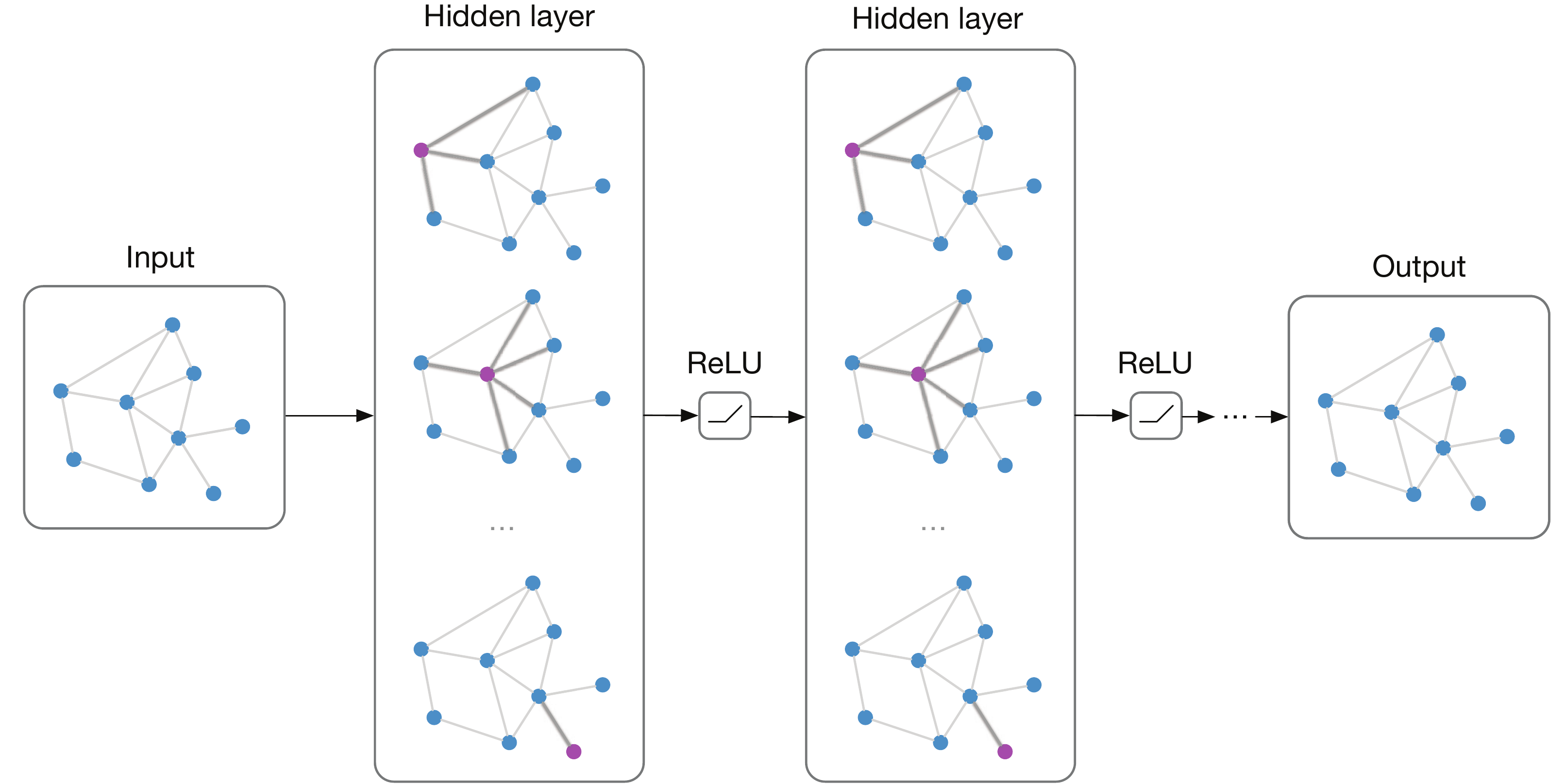}\\
\caption{An illustrative example of the spatial convolution operation proposed by Kipf and Welling~\cite{kipf2017semi} (reprinted with permission). Nodes are affected only by their immediate neighbors in each convolutional layer.}\label{fig:GCN}
\end{figure}

Kipf and Welling~\cite{kipf2017semi} further simplified the filtering by using only the first-order neighbors:
\begin{equation}\label{eq:gcn4}
   \mathbf{h}^{l+1}_i = \rho \left(\sum_{j \in \tilde{\mathcal{N}}(i)} \frac{1}{\sqrt{\tilde{\mathbf{D}}(i,i)  \tilde{\mathbf{D}}(j,j) }} \mathbf{h}^l_j \mathbf{\Theta}^l \right),
\end{equation}
where $\mathbf{h}^{l}_i \in \mathbb{R}^{f_l}$ is the hidden representation of node $v_i$ in the $l^{th}$ layer\footnote{We use a different letter because $\mathbf{h}^l\in \mathbb{R}^{f_l}$ is the hidden representation of one node, while $\mathbf{u}^l\in \mathbb{R}^{N}$ represents a dimension for all nodes.}, $\tilde{\mathbf{D}} = \mathbf{D} + \mathbf{I}$, and $\tilde{\mathcal{N}}(i) = \mathcal{N}(i) \cup \{i \}$. This can be written equivalently in an matrix form as follows:
\begin{equation}\label{eq:gcn2}
    \mathbf{H}^{l+1} = \rho\left( \tilde{\mathbf{D}}^{-\frac{1}{2}} \tilde{\mathbf{A}} \tilde{\mathbf{D}}^{-\frac{1}{2}} \mathbf{H}^{l} \mathbf{\Theta}^l \right),
\end{equation}
where $\tilde{\mathbf{A}} = \mathbf{A} + \mathbf{I}$, i.e., adding a self-connection. The authors showed that Eq.~\eqref{eq:gcn2} is a special case of Eq.~\eqref{eq:cheby1} by setting $K=1$ with a few minor changes. Then, the authors argued that stacking an adequate number of layers as illustrated in Figure~\ref{fig:GCN} has a modeling capacity similar to ChebNet but leads to better results.

An important insight of ChebNet and its extension is that they connect the spectral graph convolution with the spatial architecture. Specifically, they show that when the spectral convolution function is polynomial or first-order, the spectral graph convolution is equivalent to a spatial convolution. In addition, the convolution in Eq.~\eqref{eq:gcn4} is highly similar to the state definition in a GNN in Eq.~\eqref{eq:gnn1}, except that the convolution definition replaces the recursive definition. From this aspect, a GNN can be regarded as a GCN with a large number of identical layers to reach stable states~\cite{bronstein2017geometric}, i.e., a GNN uses a fixed function with fixed parameters to iteratively update the node hidden states until reaching an equilibrium, while a GCN has a preset number of layers and each layer contains different parameters.

Some spectral methods have also been proposed to solve the efficiency problem. For example, instead of using the Chebyshev expansion as in Eq.~\eqref{eq:cheby3}, CayleyNet~\cite{levie2017cayleynets} adopted Cayley polynomials to define graph convolutions:
\begin{equation}
    \mathbf{\Theta}(\mathbf{\Lambda}) = \theta_0  + 2Re \left\{ \sum \nolimits_{k=1}^{K} \theta_k \left(\theta_h\mathbf{\Lambda} - i\mathbf{I} \right)^k \left(\theta_h\mathbf{\Lambda} +i\mathbf{I} \right)^k \right\},
\end{equation}
where $i=\sqrt{-1}$ denotes the imaginary unit and $\theta_h$ is another spectral zoom parameter. In addition to showing that CayleyNet is as efficient as ChebNet, the authors demonstrated that the Cayley polynomials can detect ``narrow frequency bands of importance'' to achieve better results. Graph wavelet neural network (GWNN)~\cite{xu2019graph} was further proposed to replace the Fourier transform in spectral filters by the graph wavelet transform by rewriting Eq.~\eqref{eq:spectral} as follows:
\begin{equation}
    \mathbf{u}_1 *_G \mathbf{u}_2 = \mathbf{\psi} \left( \left( \mathbf{\psi}^{-1} \mathbf{u}_1 \right) \odot \left( \mathbf{\psi}^{-1} \mathbf{u}_2 \right) \right),
\end{equation}
where $\mathbf{\psi}$ denotes the graph wavelet bases. By using fast approximating algorithms to calculate $\mathbf{\psi}$ and $\mathbf{\psi}^{-1}$, GWNN's computational complexity is also $O(KM)$, i.e., linear with respect to the number of edges.
\subsubsection{The Aspect of Multiple Graphs}\label{sec:mg}
A parallel series of works has focuses on generalizing graph convolutions to multiple graphs of arbitrary sizes. Neural FPs~\cite{duvenaud2015convolutional} proposed a spatial method that also used the first-order neighbors:
\begin{equation}\label{eq:gcn3}
    \mathbf{h}^{l+1}_i = \sigma\left( \sum \nolimits_{j \in \hat{\mathcal{N}}(i)} \mathbf{h}^{l}_j \mathbf{\Theta}^l \right).
\end{equation}
Because the parameters $\mathbf{\Theta}$ can be shared across different graphs and are independent of the graph size, Neural FPs can handle multiple graphs of arbitrary sizes. Note that Eq.~\eqref{eq:gcn3} is very similar to Eq.~\eqref{eq:gcn4}. However, instead of considering the influence of node degree by adding a normalization term, Neural FPs proposed learning different parameters $\mathbf{\Theta}$ for nodes with different degrees. This strategy performed well for small graphs such as molecular graphs (i.e., atoms as nodes and bonds as edges), but may not be scalable to larger graphs.

PATCHY-SAN~\cite{niepert2016learning} adopted a different idea. It assigned a unique node order using a graph labeling procedure such as the Weisfeiler-Lehman kernel~\cite{shervashidze2011weisfeiler} and then arranged node neighbors in a line using this pre-defined order. In addition, PATCHY-SAN defined a ``receptive field'' for each node $v_i$ by selecting a fixed number of nodes from its $k$-step neighborhoods $\mathcal{N}_k(i)$. Then a standard 1-D CNN with proper normalization was adopted. Using this approach, nodes in different graphs all have a ``receptive field'' with a fixed size and order; thus, PATCHY-SAN can learn from multiple graphs like normal CNNs learn from multiple images. The drawbacks are that the convolution depends heavily on the graph labeling procedure which is a preprocessing step that is not learned.
LGCN~\cite{gao2018large} further proposed to simplify the sorting process by using a lexicographical order (i.e., sorting neighbors based on their hidden representation in the final layer $\mathbf{H}^{L}$). Instead of using a single order, the authors sorted different channels of $\mathbf{H}^{L}$ separately. SortPooling~\cite{zhang2018end} took a similar approach, but rather than sorting the neighbors of each node, the authors proposed to sort all the nodes (i.e., using a single order for all the neighborhoods). Despite the differences among these methods, enforcing a 1-D node order may not be a natural choice for graphs.

DCNN~\cite{atwood2016diffusion} adopted another approach by replacing the eigenbasis of the graph convolution with a diffusion-basis, i.e., the neighborhoods of nodes were determined by the diffusion transition probability between nodes. Specifically, the convolution was defined as follows:
\begin{equation}
    \mathbf{H}^{l+1} = \rho \left( \mathbf{P}^K \mathbf{H}^l \mathbf{\Theta}^l \right),
\end{equation}
where $\mathbf{P}^K = \left( \mathbf{P} \right)^K$ is the transition probability of a length-$K$ diffusion process (i.e., random walks), $K$ is a preset diffusion length, and $\mathbf{\Theta}^l$ are learnable parameters. Because only $\mathbf{P}^K$ depends on the graph structure, the parameters $\mathbf{\Theta}^l$ can be shared across graphs of arbitrary sizes. However, calculating $\mathbf{P}^K$ has a time complexity of $O\left(N^2K\right)$; thus, this method is not scalable to large graphs.

DGCN~\cite{zhuang2018dual} was further proposed to jointly adopt the diffusion and the adjacency bases using a dual graph convolutional network. Specifically, DGCN used two convolutions: one was Eq.~\eqref{eq:gcn2}, and the other replaced the adjacency matrix with the positive pointwise mutual information (PPMI) matrix~\cite{levy2014neural} of the transition probability as follows:
\begin{equation}
    \mathbf{Z}^{l+1} = \rho\left(\mathbf{D}^{-\frac{1}{2}}_{P} \mathbf{X}_{P} \mathbf{D}^{-\frac{1}{2}}_{P} \mathbf{Z}^{l} \mathbf{\Theta}^l \right),
\end{equation}
where $\mathbf{X}_{P}$ is the PPMI matrix calculated as:
\begin{equation}\label{eq:PPMI}
    \mathbf{X}_{P}(i,j) = \max\left(\log\left(\frac{\mathbf{P}(i,j)\sum_{i,j}\mathbf{P}(i,j)}{\sum_i \mathbf{P}(i,j) \sum_j \mathbf{P}(i,j)}\right),0\right),
\end{equation}
and $\mathbf{D}_{P}(i,i) = \sum_{j}\mathbf{X}_{P}(i,j)$ is the diagonal degree matrix of $\mathbf{X}_P$. Then, these two convolutions were ensembled by minimizing the mean square differences between $\mathbf{H}$ and $\mathbf{Z}$. DGCN adopted a random walk sampling technique to accelerate the transition probability calculation. The experiments demonstrated that such dual convolutions were effective even for single-graph problems.

\subsubsection{Frameworks}\label{sec:framework}
Based on the above two lines of works, MPNNs~\cite{gilmer2017neural} were proposed as a unified framework for the graph convolution operation in the spatial domain using message-passing functions:
\begin{equation}\label{eq:gcn6}
\begin{gathered}
    \mathbf{m}_i^{l+1} = \sum \nolimits_{j \in \mathcal{N}(i)} \mathcal{F}^l\left(  \mathbf{h}_i^l,\mathbf{h}_j^l,\mathbf{F}^E_{i,j}\right) \\
    \mathbf{h}_i^{l+1} = \mathcal{G}^l\left( \mathbf{h}_i^l,\mathbf{m}_i^{l+1}\right),
\end{gathered}
\end{equation}
where $\mathcal{F}^l(\cdot)$ and $\mathcal{G}^l(\cdot)$ are the message functions and vertex update functions to be learned, respectively, and $\mathbf{m}^l$ denotes the ``messages'' passed between nodes. Conceptually, MPNNs are a framework in which each node sends messages based on its states and updates its states based on messages received from the immediate neighbors. The authors showed that the above framework had included many existing methods such as GGS-NNs~\cite{li2016gated}, Bruna~\textit{et~al.}~\cite{bruna2014spectral}, Henaff~\textit{et~al.}~\cite{henaff2015deep}, Neural FPs~\cite{duvenaud2015convolutional}, Kipf and Welling~\cite{kipf2017semi} and Kearnes~\textit{et~al.}~\cite{kearnes2016molecular} as special cases. In addition, the authors proposed adding a ``master'' node that was connected to all the nodes to accelerate the message-passing across long distances, and they split the hidden representations into different ``towers'' to improve the generalization ability. The authors showed that a specific variant of MPNNs could achieve state-of-the-art performance in predicting molecular properties.

Concurrently, GraphSAGE~\cite{hamilton2017inductive} took a similar idea as Eq.~\eqref{eq:gcn6} using multiple aggregating functions as follows:
\begin{equation}\label{eq:gcn7}
\begin{gathered}
    \mathbf{m}_i^{l+1} = \text{AGGREGATE}^l(\{ \mathbf{h}_j^l, \forall j \in \mathcal{N}(i) \}) \\
    \mathbf{h}_i^{l+1} = \rho \left( \mathbf{\Theta}^l \left[\mathbf{h}_i^l,\mathbf{m}_i^{l+1} \right]\right),
\end{gathered}
\end{equation}
where $\left[\cdot,\cdot \right]$ is the concatenation operation and $\text{AGGREGATE}(\cdot)$ represents the aggregating function. The authors suggested three aggregating functions: the element-wise mean, an LSTM, and max-pooling as follows:
\begin{equation}
    \text{AGGREGATE}^l = \max \{\rho(\mathbf{\Theta}_{\text{pool}} \mathbf{h}_j^l + \mathbf{b}_{\text{pool}}), \forall j \in \mathcal{N}(i)  \} ,
\end{equation}
where $\mathbf{\Theta}_{\text{pool}}$ and $\mathbf{b}_{\text{pool}}$ are the parameters to be learned and $\max\left\{ \cdot \right\}$ is the element-wise maximum. For the LSTM aggregating function, because an neighbors order is needed, the authors adopted a simple random order.

Mixture model network (MoNet)~\cite{monti2017geometric} also tried to unify the existing GCN models as well as CNNs for manifolds into a common framework using ``template matching'':
\begin{equation}\label{eq:gcn8}
    h^{l+1}_{ik} = \sum \nolimits_{j \in \mathcal{N}(i)} \mathcal{F}^l_k(\mathbf{u}(i,j)) \mathbf{h}^l_j, k = 1,...,f_{l+1},
\end{equation}
where $\mathbf{u}(i,j)$ are the pseudo-coordinates of the node pair $(v_i,v_j)$, $\mathcal{F}^l_k(\mathbf{u})$ is a parametric function to be learned, and $h^{l}_{ik}$ is the $k^{th}$ dimension of $\mathbf{h}^l_i$. In other words, $\mathcal{F}^l_k(\mathbf{u})$ served as a weighting kernel for combining neighborhoods. Then, MoNet adopted the following Gaussian kernel:
\begin{equation}
    \mathcal{F}^l_k(\mathbf{u}) = \exp\left( -\frac{1}{2} (\mathbf{u} - \boldsymbol\mu^l_k)^T (\mathbf{\Sigma}^l_k)^{-1} (\mathbf{u} - \boldsymbol\mu^l_k )\right),
\end{equation}
where $\boldsymbol\mu^l_k$ and $\mathbf{\Sigma}^l_k$ are the mean vectors and diagonal covariance matrices to be learned, respectively. The pseudo-coordinates were degrees as in Kipf and Welling~\cite{kipf2017semi}, i.e.,
\begin{equation}
\mathbf{u}(i,j) = (\frac{1}{\sqrt{\mathbf{D}(i,i)}}, \frac{1}{\sqrt{\mathbf{D}(j,j)}}).
\end{equation}

Graph networks (GNs)~\cite{battaglia2018relational} proposed a more general framework for both GCNs and GNNs that learned three sets of representations: $\mathbf{h}_i^l,\mathbf{e}_{ij}^l$, and $\mathbf{z}^l$ as the representation for nodes, edges, and the entire graph, respectively. These representations were learned using three aggregation and three updating functions:
\begin{equation}\label{eq:gcn10}
\begin{gathered}
    \mathbf{m}_i^l = \mathcal{G}^{E \rightarrow V}(\{ \mathbf{e}_{ij}^l, \forall j \in \mathcal{N}(i) \}), \mathbf{m}_V^l = \mathcal{G}^{V \rightarrow G}( \{ \mathbf{h}_i^l, \forall v_i \in V \} ) \\
    \mathbf{m}_E^l = \mathcal{G}^{E \rightarrow G}( \{ \mathbf{e}_{ij}^l, \forall (v_i,v_j) \in E \}), \mathbf{h}_i^{l+1} = \mathcal{F}^V(\mathbf{m}_i^l,\mathbf{h}_i^l,\mathbf{z}^l) \\
    \mathbf{e}_{ij}^{l+1} = \mathcal{F}^E(\mathbf{e}_{ij}^l,\mathbf{h}_i^l,\mathbf{h}_j^l,\mathbf{z}^l), \mathbf{z}^{l+1} = \mathcal{F}^G(\mathbf{m}_E^l,\mathbf{m}_V^l,\mathbf{z}^l),
\end{gathered}
\end{equation}
where $\mathcal{F}^V(\cdot),\mathcal{F}^E(\cdot)$, and $\mathcal{F}^G(\cdot)$ are the corresponding updating functions for nodes, edges, and the entire graph, respectively, and $\mathcal{G}(\cdot)$ represents message-passing functions whose superscripts denote message-passing directions. Note that the message-passing functions all take a set as the input, thus their arguments are variable in length and these functions should be invariant to input permutations; some examples include the element-wise summation, mean, and maximum. Compared with MPNNs, GNs introduced the edge representations and the representation of the entire graph, thus making the framework more general.

In summary, the convolution operations have evolved from the spectral domain to the spatial domain and from multistep neighbors to the immediate neighbors. Currently, gathering information from the immediate neighbors (as in Eq.~\eqref{eq:gcn2}) and following the framework of Eqs.~\eqref{eq:gcn6}\eqref{eq:gcn7}\eqref{eq:gcn10} are the most common choices for graph convolution operations.

\subsection{Readout Operations}
Using graph convolution operations, useful node features can be learned to solve many node-focused tasks. However, to tackle graph-focused tasks, node information needs to be aggregated to form a graph-level representation. In the literature, such procedures are usually called the readout operations\footnote{Readout operations are also related to graph coarsening, i.e., reducing a large graph to a smaller graph, because a graph-level representation can be obtained by coarsening the graph to a single node. Some papers use these two terms interchangeably.}.
Based on a regular and local neighborhood, standard CNNs conduct multiple stride convolutions or poolings to gradually reduce the resolution. Since graphs lack a grid structure, these existing methods cannot be used directly.

\textbf{Order invariance}. A critical requirement for the graph readout operations is that the operation should be invariant to the node order, i.e., if we change the indices of nodes and edges using a bijective function between two node sets, the representation of the entire graph should not change. For example, whether a drug can treat certain diseases depends on its inherent structure; thus, we should get identical results if we represent the drug using different node indices. Note that because this problem is related to the graph isomorphism problem, of which the best-known algorithm is quasipolynomial~\cite{babai2016graph}, we only can find a function that is order-invariant but not vice versa in a polynomial time, i.e., even two structurally different graphs may have the same representation.

\subsubsection{Statistics}
The most basic order-invariant operations involve simple statistics such as summation, averaging or max-pooling~\cite{duvenaud2015convolutional,atwood2016diffusion}, i.e.,
\begin{equation}\label{eq:gcnr1}
    \mathbf{h}_G = \sum_{i=1}^N \mathbf{h}^L_i \; \text{or} \; \mathbf{h}_G = \frac{1}{N}\sum_{i=1}^N \mathbf{h}^L_i \; \text{or} \; \mathbf{h}_G = \max\left\{\mathbf{h}^L_i,\forall i \right\},
\end{equation}
where $\mathbf{h}_G$ is the representation of the graph $G$ and $\mathbf{h}^L_i$ is the representation of node $v_i$ in the final layer $L$. However, such first-moment statistics may not be sufficiently representative to distinguish different graphs.

Kearnes~\textit{et~al.}~\cite{kearnes2016molecular} suggested considering the distribution of node representations by using fuzzy histograms~\cite{klir1995fuzzy}. The basic idea behind fuzzy histograms is to construct several ``histogram bins'' and then calculate the memberships of $\mathbf{h}^L_i$ to these bins, i.e., by regarding node representations as samples and matching them to some pre-defined templates, and finally return the concatenation of the final histograms. In this way, nodes with the same sum/average/maximum but with different distributions can be distinguished.

Another commonly used approach for aggregating node representation is to add a fully connected (FC) layer as the final layer~\cite{bruna2014spectral}, i.e.,
\begin{equation}\label{eq:gcnr2}
    \mathbf{h}_G =  \rho \left( \left[\mathbf{H}^L\right] \mathbf{\Theta}_{FC} \right),
\end{equation}
where $\left[\mathbf{H}^L\right]\in \mathbb{R}^{Nf_L}$ is the concatenation of the final node representation $\mathbf{H}^L$, $\mathbf{\Theta}_{FC} \in \mathbb{R}^{Nf_L \times f_{\text{output}}}$ are parameters, and $f_{\text{output}}$ is the dimensionality of the output. Eq.~\eqref{eq:gcnr2} can be regarded as a weighted sum of node-level features. One advantage is that the model can learn different weights for different nodes; however, this ability comes at the cost of being unable to guarantee order invariance.

\subsubsection{Hierarchical Clustering}
\begin{figure}
\centering
\includegraphics[width = 8.5cm]{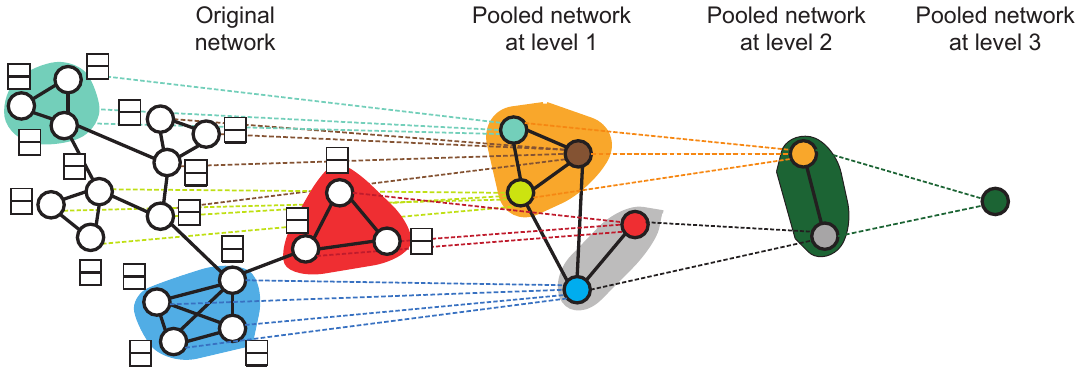}\\
\caption{An example of performing a hierarchical clustering algorithm. Reprinted from~\cite{ying2018hierarchical} with permission.}\label{fig:HC}
\end{figure}

Rather than a dichotomy between node and graph level structures, graphs are known to exhibit rich hierarchical structures~\cite{ma2018hierarchical}, which can be explored by hierarchical clustering methods as shown in Figure~\ref{fig:HC}. For example, a density-based agglomerative clustering~\cite{Ruppert2010The} was used in Bruna~\textit{et~al.}~\cite{bruna2014spectral} and multi-resolution spectral clustering~\cite{von2007tutorial} was used in Henaff~\textit{et~al.}~\cite{henaff2015deep}. ChebNet~\cite{defferrard2016convolutional} and MoNet~\cite{monti2017geometric} adopted another greedy hierarchical clustering algorithm, Graclus~\cite{dhillon2007weighted}, to merge two nodes at a time, along with a fast pooling method to rearrange the nodes into a balanced binary tree. ECC~\cite{simonovsky2017dynamic} adopted another hierarchical clustering method by performing eigendecomposition~\cite{shuman2016multiscale}. However, these hierarchical clustering methods are all independent of the graph convolutions (i.e., they can be performed as a preprocessing step and are not trained in an end-to-end fashion).

To solve that problem, DiffPool~\cite{ying2018hierarchical} proposed a differentiable hierarchical clustering algorithm jointly trained with the graph convolutions. Specifically, the authors proposed learning a soft cluster assignment matrix in each layer using the hidden representations as follows:
\begin{equation}
    \mathbf{S}^l = \mathcal{F}\left(\mathbf{A}^l,\mathbf{H}^l \right),
\end{equation}
where $\mathbf{S}^l \in \mathbb{R}^{N_l \times N_{l+1}}$ is the cluster assignment matrix, $N_l$ is the number of clusters in the layer $l$ and $\mathcal{F}(\cdot)$ is a function to be learned. Then, the node representations and the new adjacency matrix for this ``coarsened'' graph can be obtained by taking the average according to $\mathbf{S}^l$ as follows:
\begin{equation}
    \mathbf{H}^{l+1} = (\mathbf{S}^l)^T \hat{\mathbf{H}}^{l+1},
    \mathbf{A}^{l+1} = (\mathbf{S}^l)^T \mathbf{A}^l \mathbf{S}^l,
\end{equation}
where $\hat{\mathbf{H}}^{l+1}$ is obtained by applying a graph convolution layer to $\mathbf{H}^l$, i.e., coarsening the graph from $N_l$ nodes to $N_{l+1}$ nodes in each layer after the convolution operation. The initial number of nodes is $N_0=N$ and the last layer is $N_L=1$, i.e., a single node that represents the entire graph. Because the cluster assignment operation is soft, the connections between clusters are not sparse; thus the time complexity of the method is $O(N^2)$ in principle.

\subsubsection{Imposing Orders and Others}
As mentioned in Section~\ref{sec:mg}, PATCHY-SAN~\cite{niepert2016learning} and SortPooling~\cite{zhang2018end} took the idea of imposing a node order and then resorted to standard 1-D pooling as in CNNs. Whether these methods can preserve order invariance depends on how the order is imposed, which is another research field that we refer readers to~\cite{Mckay2014Practical} for a survey. However, whether imposing a node order is a natural choice for graphs and if so, what the best node orders are constitute still on-going research topics.

In addition to the aforementioned methods, there are some heuristics. In GNNs~\cite{scarselli2009graph}, the authors suggested adding a special node connected to all nodes to represent the entire graph. Similarly, GNs~\cite{battaglia2018relational} proposed to directly learn the representation of the entire graph by receiving messages from all nodes and edges.

MPNNs adopted set2set~\cite{vinyals2016order}, a modification of the seq2seq model. Specifically, set2set uses a ``Read-Process-and-Write'' model that receives all inputs simultaneously, computes internal memories using an attention mechanism and an LSTM, and then writes the outputs. Unlike seq2seq which is order-sensitive, set2set is invariant to the input order.

\subsubsection{Summary}
In short, statistics such as averaging or summation are the most simple readout operations, while hierarchical clustering algorithms jointly trained with graph convolutions are more advanced but are also more sophisticated. Other methods such as adding a pseudo node or imposing a node order have also been investigated.

\subsection{Improvements and Discussions}
Many techniques have been introduced to further improve GCNs. Note that some of these methods are general and could be applied to other deep learning models on graphs as well.

\subsubsection{Attention Mechanism}
\begin{figure}
\centering
\includegraphics[width = 5cm]{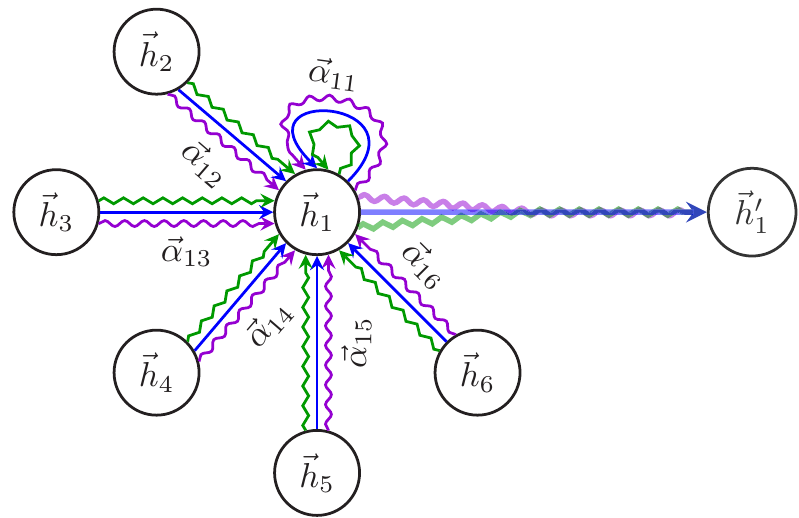}\\
\caption{An illustration of the multi-head attention mechanism proposed in GAT~\cite{velickovic2018graph} (reprinted with permission). Each color denotes an independent attention vector.}\label{fig:attention}
\end{figure}

In the aforementioned GCNs, the node neighborhoods are aggregated with equal or pre-defined weights. However, the influences of neighbors can vary greatly; thus, they should be learned during training rather than being predetermined. Inspired by the attention mechanism~\cite{vaswani2017attention}, graph attention network (GAT)~\cite{velickovic2018graph} introduces the attention mechanism into GCNs by modifying the convolution operation in Eq.~\eqref{eq:gcn4} as follows:
\begin{equation}\label{eq:gcn5}
   \mathbf{h}^{l+1}_i = \rho \left(\sum \nolimits_{j \in \hat{\mathcal{N}}(i)} \alpha_{ij}^l \mathbf{h}^l_j \mathbf{\Theta}^l \right),
\end{equation}
where $\alpha_{ij}^l$ is node $v_i$'s attention to node $v_j$ in the $l^{th}$ layer:
\begin{equation}
    \alpha_{ij}^l = \frac{\exp \left(\text{LeakyReLU} \left( \mathcal{F} \left( \mathbf{h}^l_i \mathbf{\Theta}^l, \mathbf{h}^l_j \mathbf{\Theta}^l\right) \right) \right)}{\sum_{k \in \hat{\mathcal{N}}(i)} \exp\left(\text{LeakyReLU}  \left( \mathcal{F}\left( \mathbf{h}^l_i \mathbf{\Theta}^l , \mathbf{h}^l_k \mathbf{\Theta}^l\right) \right) \right)},
\end{equation}
where $\mathcal{F}(\cdot,\cdot)$ is another function to be learned such as a multi-layer perceptron (MLP). To improve model capacity and stability, the authors also suggested using multiple independent attentions and concatenating the results, i.e., the multi-head attention mechanism~\cite{vaswani2017attention} as illustrated in Figure~\ref{fig:attention}.
GaAN~\cite{zhang2018gaan} further proposed learning different weights for different heads and applied such a method to the traffic forecasting problem.

HAN~\cite{wang2019hetero} proposed a two-level attention mechanism, i.e., a node-level and a semantic-level attention mechanism, for heterogeneous graphs. Specifically, the node-level attention mechanism was similar to a GAT, but also considerd node types; therefore, it could assign different weights to aggregating meta-path-based neighbors. The semantic-level attention then learned the importance of different meta-paths and outputed the final results.

\subsubsection{Residual and Jumping Connections}
Many researches have observed that the most suitable depth for the existing GCNs is often very limited, e.g., 2 or 3 layers. This problem is potentially due to the practical difficulties involved in training deep GCNs or the over-smoothing problem, i.e., all nodes in deeper layers have the same representation~\cite{li2018deeper,xu2018representation}. To remedy this problem, residual connections similar to ResNet~\cite{he2016deep} can be added to GCNs. For example, Kipf and Welling~\cite{kipf2017semi} added residual connections into Eq.~\eqref{eq:gcn2} as follows:
\begin{equation}
    \mathbf{H}^{l+1} = \rho\left( \tilde{\mathbf{D}}^{-\frac{1}{2}} \tilde{\mathbf{A}} \tilde{\mathbf{D}}^{-\frac{1}{2}} \mathbf{H}^{l} \mathbf{\Theta}^l \right) + \mathbf{H}^{l}.
\end{equation}
They showed experimentally that adding such residual connections could allow the depth of the network to increase, which is similar to the results of ResNet.

Column network (CLN)~\cite{pham2017column} adopted a similar idea by using the following residual connections with learnable weights:
\begin{equation}\label{eq:gcn9}
    \mathbf{h}^{l+1}_i = \boldsymbol\alpha^l_i \odot \widetilde{\mathbf{h}}^{l+1}_i  + (1 - \boldsymbol\alpha^l_i) \odot \mathbf{h}^l_i,
\end{equation}
where $\widetilde{\mathbf{h}}^{l+1}_i$ is calculated similar to Eq.~\eqref{eq:gcn2} and $\boldsymbol\alpha^l_i$ is a set of weights calculated as follows:
\begin{equation}
     \boldsymbol\alpha^l_i = \rho \left( \mathbf{b}^l_\alpha + \mathbf{\Theta}_\alpha^l \mathbf{h}_i^l + \mathbf{\Theta}_\alpha^{'l} \sum \nolimits_{j \in \mathcal{N}(i)} \mathbf{h}_j^l \right),
\end{equation}
where $\mathbf{b}^l_\alpha, \mathbf{\Theta}_\alpha^l,\mathbf{\Theta}_\alpha^{'l}$ are parameters. Note that Eq.~\eqref{eq:gcn9} is very similar to the GRU as in GGS-NNs~\cite{li2016gated}. The differences are that in a CLN, the superscripts denote the number of layers, and different layers contain different parameters, while in GGS-NNs, the superscript denotes the pseudo time and a single set of parameters is used across time steps.

Inspired by personalized PageRank, PPNP~\cite{klicpera2019predict} defined graph convolutions with teleportation to the initial layer:
\begin{equation}
    \mathbf{H}^{l+1} = (1 - \alpha) \tilde{\mathbf{D}}^{-\frac{1}{2}} \tilde{\mathbf{A}} \tilde{\mathbf{D}}^{-\frac{1}{2}} \mathbf{H}^{l} + \alpha \mathbf{H}^{0},
\end{equation}
where $\mathbf{H}_0 = \mathcal{F}_{\theta}(\mathbf{F}^V)$ and $\alpha$ is a hyper-parameter. Note that all the parameters are in $\mathcal{F}_{\theta}(\cdot)$ rather than in the graph convolutions.

\begin{figure}
\centering
\includegraphics[width = 5cm]{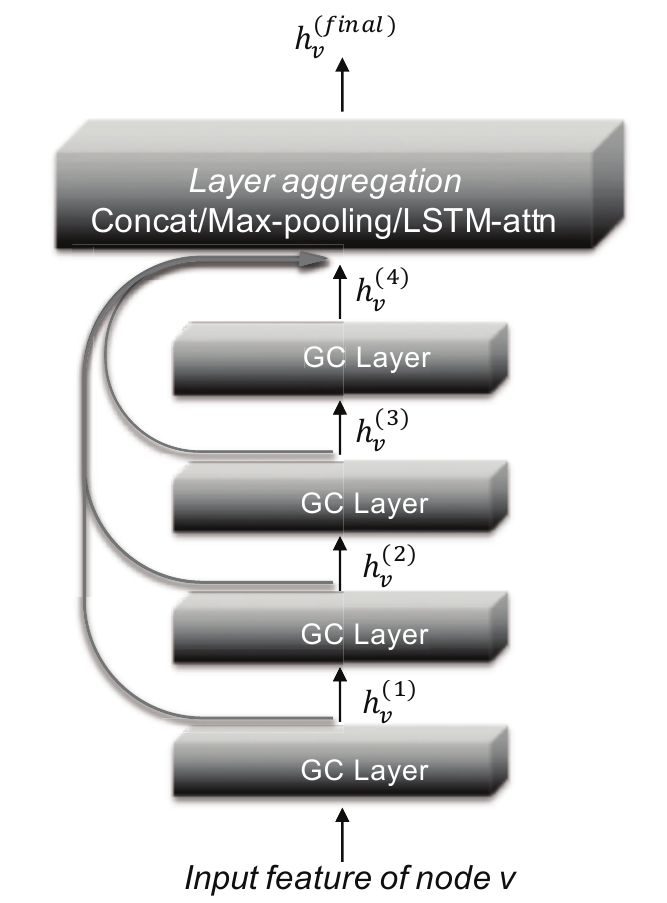}\\
\caption{Jumping knowledge networks proposed in~\cite{xu2018representation} in which the last layer is connected to all the other layers to selectively exploit different information from different layers. GC denotes graph convolutions. Reprinted with permission.}\label{fig:JK}
\end{figure}

Jumping knowledge networks (JK-Nets)~\cite{xu2018representation} proposed another architecture to connect the last layer of the network with all the lower hidden layers, i.e., by ``jumping'' all the representations to the final output, as illustrated in Figure~\ref{fig:JK}. In this way, the model can learn to selectively exploit information from different layers. Formally, JK-Nets was formulated as follows:
\begin{equation}
    \mathbf{h}_i^{\text{final}} = \text{AGGREGATE}(\mathbf{h}_i^0,\mathbf{h}_i^1,...,\mathbf{h}_i^L),
\end{equation}
where $\mathbf{h}_i^{\text{final}}$ is the final representation for node $v_i$, AGGREGATE$(\cdot)$ is the aggregating function, and $L$ is the number of hidden layers. JK-Nets used three aggregating functions similar to GraphSAGE~\cite{hamilton2017inductive}: concatenation, max-pooling, and the LSTM attention. The experimental results showed that adding jump connections could improve the performance of multiple GCNs.

\subsubsection{Edge Features}
The aforementioned GCNs mostly focus on utilizing node features and graph structures. In this subsection, we briefly discuss how to use another important source of information: the edge features.

For simple edge features with discrete values such as the edge type, a straightforward method is to train different parameters for different edge types and aggregate the results. For example, Neural FPs~\cite{duvenaud2015convolutional} trained different parameters for nodes with different degrees, which corresponds to the implicit edge feature of bond types in a molecular graph, and then summed over the results. CLN~\cite{pham2017column} trained different parameters for different edge types in a heterogeneous graph and averaged the results. Edge-conditioned convolution (ECC)~\cite{simonovsky2017dynamic} also trained different parameters based on edge types and applied them to graph classification. Relational GCNs (R-GCNs)~\cite{schlichtkrull2018modeling} adopted a similar idea for knowledge graphs by training different weights for different relation types. However, these methods are suitable only for a limited number of discrete edge features.

DCNN~\cite{atwood2016diffusion} proposed another method to convert each edge into a node connected to the head and tail node of that edge. After this conversion, edge features can be treated as node features.

LGCN~\cite{chen2019supervised} constructed a line graph $\mathbf{B} \in \mathbb{R}^{2M \times 2M}$ to incorporate edge features as follows:
\begin{equation}
    \mathbf{B}_{i\rightarrow j, i' \rightarrow j'} = \left\{
        \begin{aligned}
            & 1 \quad \text{if} \; j=i' \; \text{and} \; j' \neq i,\\
            & 0 \quad \text{otherwise}.
        \end{aligned}
        \right.
\end{equation}
In other words, nodes in the line graph are directed edges in the original graph, and two nodes in the line graph are connected if information can flow through their corresponding edges in the original graph. Then, LGCN adopted two GCNs: one on the original graph and one on the line graph.

Kearnes~\textit{et~al.}~\cite{kearnes2016molecular} proposed an architecture using a ``weave module''. Specifically, they learned representations for both nodes and edges and exchanged information between them in each weave module using four different functions: node-to-node (NN), node-to-edge (NE), edge-to-edge (EE) and edge-to-node (EN):
\begin{equation}
\begin{gathered}
    \mathbf{h}^{l'}_i = \mathcal{F}_{NN}(\mathbf{h}^0_i,\mathbf{h}^1_i,...,\mathbf{h}^l_i), \mathbf{h}^{l''}_i = \mathcal{F}_{EN}(\{ \mathbf{e}^l_{ij} | j \in \mathcal{N}(i)\}) \\
    \mathbf{e}^{l'}_{ij} = \mathcal{F}_{EE}(\mathbf{e}^0_{ij},\mathbf{e}^1_{ij},...,\mathbf{e}^l_{ij}), \mathbf{e}^{l''}_{ij} = \mathcal{F}_{NE}(\mathbf{h}^l_i,\mathbf{h}^l_j) \\
    \mathbf{h}^{l+1}_i = \mathcal{F}_{NN}(\mathbf{h}^{l'}_i,\mathbf{h}^{l''}_i), \mathbf{e}^{l+1}_{ij} = \mathcal{F}_{EE}(\mathbf{e}^{l'}_{ij},\mathbf{e}^{l''}_{ij}),
\end{gathered}
\end{equation}
where $\mathbf{e}_{ij}^l$ is the representation of edge $(v_i,v_j)$ in the $l^{th}$ layer and $\mathcal{F}(\cdot)$ are learnable functions whose subscripts represent message-passing directions. By stacking multiple such modules, information can propagate by alternately passing between node and edge representations. Note that in the node-to-node and edge-to-edge functions, jump connections similar to those in JK-Nets~\cite{xu2018representation} are implicitly added. GNs~\cite{battaglia2018relational} also proposed learning an edge representation and updating both node and edge representations using message-passing functions as shown in Eq.~\eqref{eq:gcn10} in Section~\ref{sec:framework}. In this aspect, the ``weave module'' is a special case of GNs that does not a representation of the entire graph.

\subsubsection{Sampling Methods}
\begin{figure}
\centering
\includegraphics[width = 8.1cm]{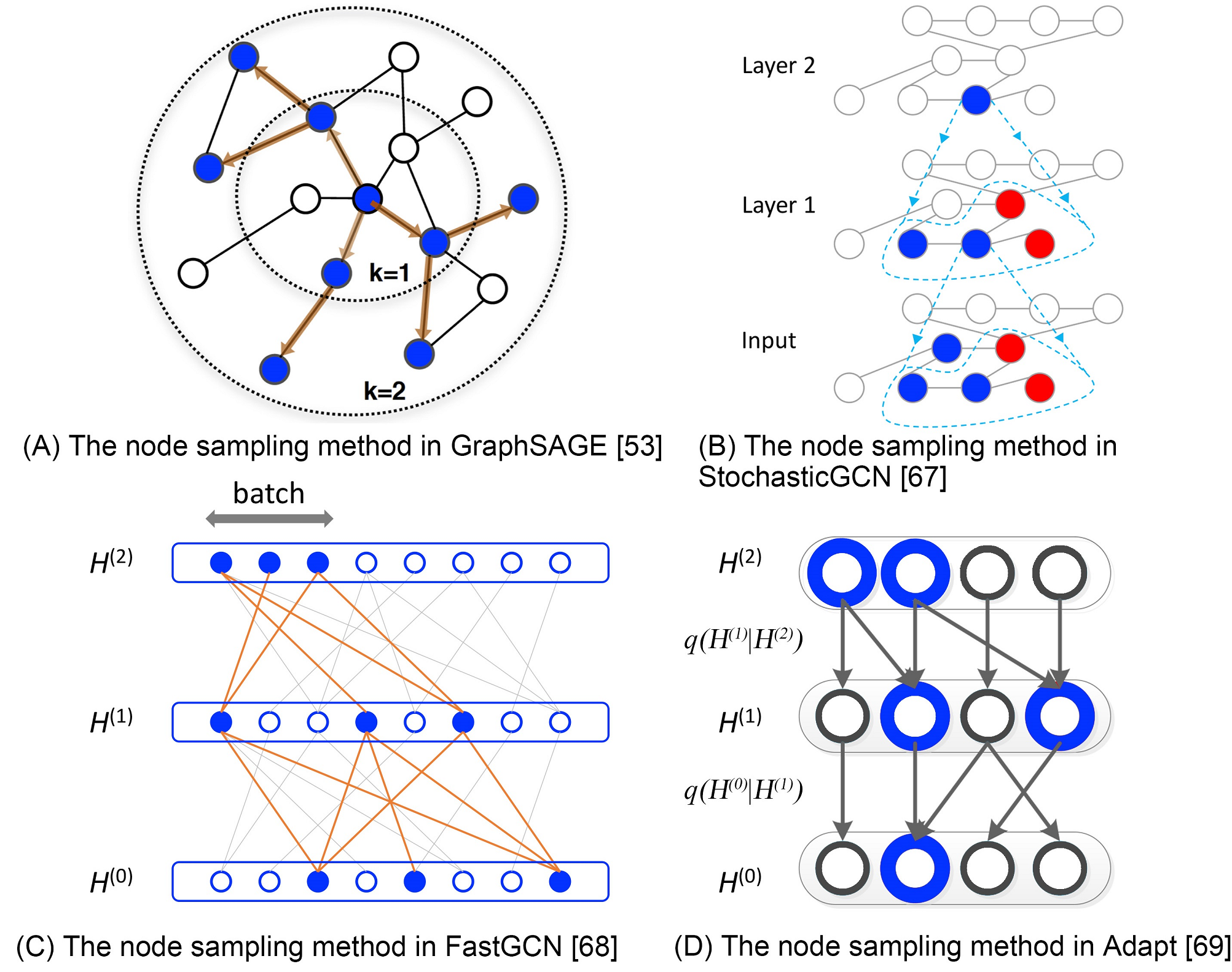}
\caption{Different node sampling methods, in which the blue nodes indicate samples from one batch and the arrows indicate the sampling directions. The red nodes in (B) represent historical samples.}\label{fig:sample}
\end{figure}

One critical bottleneck when training GCNs for large-scale graphs is efficiency. As shown in Section~\ref{sec:framework}, many GCNs follow a neighborhood aggregation scheme. However, because many real graphs follow a power-law distribution~\cite{barabasi1999emergence} (i.e., a few nodes have very large degrees), the
number of neighbors can expand extremely quickly. To deal with this problem, two types of sampling methods have been proposed: neighborhood samplings and layer-wise samplings, as illustrated in Figure~\ref{fig:sample}.

In neighborhood samplings, the sampling is performed for each node during the calculations. GraphSAGE~\cite{hamilton2017inductive} uniformly sampled a fixed number of neighbors for each node during training. PinSage~\cite{ying2018graph} proposed sampling neighbors using random walks on graphs along with several implementation improvements including coordination between the CPU and GPU, a map-reduce inference pipeline, and so on. PinSage was shown to be capable of handling a real billion-scale graph. StochasticGCN~\cite{chen2018stochastic} further proposed reducing the sampling variances by using the historical activations of the last batches as a control variate, allowing for arbitrarily small sample sizes with a theoretical guarantee.

Instead of sampling neighbors of nodes, FastGCN~\cite{chen2018fastgcn} adopted a different strategy: it sampled nodes in each convolutional layer (i.e., a layer-wise sampling) by interpreting the nodes as i.i.d. samples and the graph convolutions as integral transforms under probability measures. FastGCN also showed that sampling nodes via their normalized degrees could reduce variances and lead to better performance. Adapt~\cite{huang2018adaptive} further proposed sampling nodes in the lower layers conditioned on their top layer; this approach was more adaptive and applicable to explicitly reduce variances.

\subsubsection{Inductive Setting}
Another important aspect of GCNs is that whether they can be applied to an inductive setting, i.e., training on a set of nodes or graphs and testing on another unseen set of nodes or graphs. In principle, this goal is achieved by learning a mapping function on the given features that are not dependent on the graph basis and can be transferred across nodes or graphs. The inductive setting was verified in GraphSAGE~\cite{hamilton2017inductive}, GAT~\cite{velickovic2018graph}, GaAN~\cite{zhang2018gaan}, and FastGCN~\cite{chen2018fastgcn}. However, the existing inductive GCNs are suitable only for graphs with explicit features. How to conduct inductive learnings for graphs without explicit features, usually called the out-of-sample problem~\cite{ma2018depthlgp}, remains largely open in the literature.

\subsubsection{Theoretical Analysis}
To understand the effectiveness of GCNs, some theoretical analyses have been proposed that can be divided into three categories: node-focused tasks, graph-focused tasks, and general analysis.

For node-focused tasks, Li~\textit{et~al.}~\cite{li2018deeper} first analyzed the performance of GCNs by using a special form of Laplacian smoothing, which makes the features of nodes in the same cluster similar. The original Laplacian smoothing operation is formulated as follows:
\begin{equation}\label{eq:lsmoothing}
    \mathbf{h}'_i = (1 - \gamma)\mathbf{h}_i + \gamma \sum \nolimits_{j \in \mathcal{N}(i)} \frac{1}{d_i} \mathbf{h}_j,
\end{equation}
where $\mathbf{h}_i$ and $\mathbf{h}'_i$ are the original and smoothed features of node $v_i$, respectively. We can see that Eq.~\eqref{eq:lsmoothing} is very similar to the graph convolution in Eq.~\eqref{eq:gcn4}. Based on this insight, Li~\textit{et~al.}~also proposed a co-training and a self-training method for GCNs.

Recently, Wu~\textit{et~al.}~\cite{wu2019simplifying} analyzed GCNs from a signal processing perspective. By regarding node features as graph signals, they showed that Eq.~\eqref{eq:gcn4} is basically a fixed low-pass filter. Using this insight, they proposed
an extremely simplified graph convolution (SGC) architecture by removing all the  nonlinearities and collapsing the learning parameters into one matrix:
\begin{equation}
    \mathbf{H}^{L} = \left(\tilde{\mathbf{D}}^{-\frac{1}{2}} \tilde{\mathbf{A}} \tilde{\mathbf{D}}^{-\frac{1}{2}}\right)^L \mathbf{F}_V \mathbf{\Theta}.
\end{equation}
The authors showed that such a ``non-deep-learning'' GCN variant achieved comparable performance to existing GCNs in many tasks. Maehara~\cite{maehara2019revisiting} enhanced this result by showing that the low-pass filtering operation did not equip GCNs with a nonlinear manifold learning ability, and further proposed GFNN model to remedy this problem by adding a MLP after the graph convolution layers.

For graph-focused tasks, Kipf and Welling~\cite{kipf2017semi} and the authors of SortPooling~\cite{zhang2018end} both considered the relationship between GCNs and graph kernels such as the Weisfeiler-Lehman (WL) kernel~\cite{shervashidze2011weisfeiler}, which is widely used in graph isomorphism tests. They showed that GCNs are conceptually a generalization of the WL kernel because both methods iteratively aggregate information from node neighbors. Xu~\textit{et~al.}~\cite{xu2019powerful} formalized this idea by proving that the WL kernel provides an upper bound for GCNs in terms of distinguishing graph structures. Based on this analysis, they proposed graph isomorphism network (GIN) and showed that a readout operation using summation and a MLP can achieve provably maximum discriminative power, i.e., the highest training accuracy in graph classification tasks.

For general analysis, Scarselli~\textit{et~al.}~\cite{scarselli2018vapnik} showed that the Vapnik-Chervonenkis dimension (VC-dim) of GCNs with different activation functions has the same scale as the existing RNNs. Chen~\textit{et~al.}~\cite{chen2019supervised} analyzed the optimization landscape of linear GCNs and showed that any local minimum is relatively close to the global minimum under certain simplifications. Verma and Zhang~\cite{verma2019stability} analyzed the algorithmic stability and generalization bound of GCNs. They showed that single-layer GCNs satisfy the strong notion of uniform stability if the largest absolute eigenvalue of the graph convolution filters is independent of the graph size.

\section{Graph Autoencoders}
The autoencoder (AE) and its variations have been widely applied in unsupervised learning tasks~\cite{Vincent2008Extracting} and are suitable for learning node representations for graphs. The implicit assumption is that graphs have an inherent, potentially nonlinear low-rank structure. In this section, we first elaborate graph autoencoders and then introduce graph variational autoencoders and other improvements. The main characteristics of GAEs are summarized in Table~\ref{tab:GAE}.

    \begin{table*}
    \centering
    \caption{A Comparison among Different Graph Autoencoders (GAEs). T.C. = Time Complexity}\label{tab:GAE}
    \begin{tabular}{ | c | c | c | c | c | c |}
    \hline
    Method & Type & Objective Function      & T.C. & Node Features  & Other Characteristics    \\ \hline
    SAE~\cite{tian2014learning}    & AE     & L2-reconstruction & $O(M)$  & No  & -      \\ \hline
    SDNE~\cite{wang2016structural} & AE     & L2-reconstruction + Laplacian eigenmaps & $O(M)$ & No &  -      \\ \hline
    DNGR~\cite{cao2016deep}        & AE     & L2-reconstruction & $O(N^2)$  & No  &   -    \\ \hline
    GC-MC~\cite{berg2017graph}     & AE     & L2-reconstruction & $O(M)$  & Yes & GCN encoder       \\ \hline
    DRNE~\cite{tu2018deep}         & AE     & Recursive reconstruction    & $O(Ns)$ & No & LSTM aggregator \\ \hline
    G2G~\cite{bojchevski2018deep}  & AE     & KL + ranking      & $O(M)$  & Yes & Nodes as distributions      \\ \hline
    VGAE~\cite{kipf2016variational}& VAE    & Pairwise reconstruction & $O(N^2)$  & Yes & GCN encoder    \\ \hline
    DVNE~\cite{zhu2018deep}        & VAE    & Wasserstein + ranking & $O(M)$ & No  & Nodes as distributions   \\ \hline
    ARGA/ARVGA~\cite{pan2018adversarially}  & AE/VAE & L2-reconstruction + GAN & $O(N^2)$ & Yes & GCN encoder    \\ \hline
    NetRA~\cite{yu2018learning}    & AE     & Recursive reconstruction + Laplacian eigenmaps + GAN & $O(M)$ & No & LSTM encoder \\ \hline
    \end{tabular}
    \end{table*}

\subsection{Autoencoders}
The use of AEs for graphs originated from sparse autoencoder (SAE)~\cite{tian2014learning}. The basic idea is that, by regarding the adjacency matrix or its variations as the raw features of nodes, AEs can be leveraged as a dimensionality reduction technique to learn low-dimensional node representations. Specifically, SAE adopted the following L2-reconstruction loss:
\begin{equation}\label{eq:gae1}
\begin{gathered}
        \min_{\mathbf{\Theta}} \mathcal{L}_2 = \sum \nolimits_{i=1}^N \left\| \mathbf{P}\left(i,:\right) - \hat{\mathbf{P}}\left(i,:\right) \right\|_2 \\
        \hat{\mathbf{P}}\left(i,:\right) = \mathcal{G}\left( \mathbf{h}_i \right), \mathbf{h}_i = \mathcal{F} \left( \mathbf{P}\left(i,: \right) \right),
\end{gathered}
\end{equation}
where $\mathbf{P}$ is the transition matrix, $\hat{\mathbf{P}}$ is the reconstructed matrix, $\mathbf{h}_i \in \mathbb{R}^d$ is the low-dimensional representation of node $v_i$, $\mathcal{F}(\cdot)$ is the encoder, $\mathcal{G}(\cdot)$ is the decoder, $d \ll N$ is the dimensionality, and $\mathbf{\Theta}$ are parameters. Both the encoder and decoder are an MLP with many hidden layers. In other words, a SAE compresses the information of $\mathbf{P}(i,:)$ into a low-dimensional vector $\mathbf{h}_i$ and then reconstructs the original feature from that vector.
Another sparsity regularization term was also added. After obtaining the low-dimensional representation $\mathbf{h}_i$, k-means~\cite{macqueen1967some} was applied for the node clustering task. The experiments prove that SAEs outperform non-deep learning baselines. However, SAE was based on an incorrect theoretical analysis.\footnote{SAE~\cite{tian2014learning} motivated the problem by analyzing the connection between spectral clustering and singular value decomposition, which is mathematically incorrect as pointed out in~\cite{zhang2018note}.} The mechanism underlying its effectiveness remained unexplained.

\begin{figure}
  \centering
  \includegraphics[width = 7.7cm]{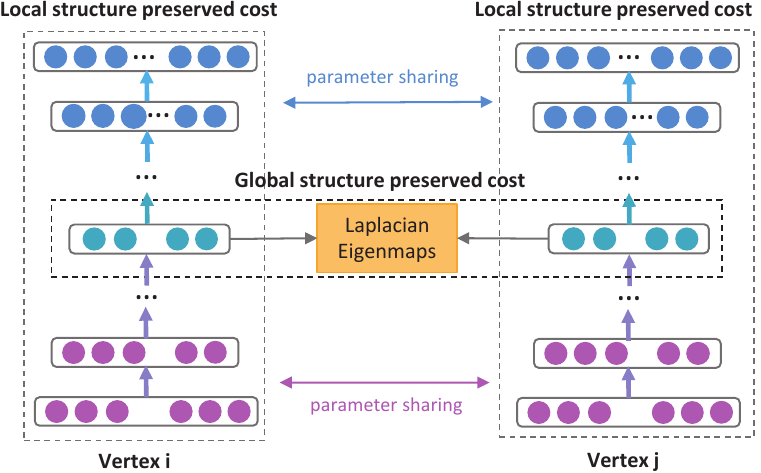}\\
  \caption{The framework of SDNE~\cite{wang2016structural}. Both the first and second-order proximities of nodes are preserved using deep autoencoders. }\label{fig:SDNE}
  \end{figure}

Structure deep network embedding (SDNE)~\cite{wang2016structural} filled in the puzzle by showing that the L2-reconstruction loss in Eq.~\eqref{eq:gae1} actually corresponds to the second-order proximity between nodes, i.e., two nodes share similar latten representations if they have similar neighborhoods, which is a well-studied concept in network science known as collaborative filtering or triangle closure~\cite{barabasi2016network}. Motivated by network embedding methods showing that the first-order proximity is also important~\cite{tang2015line}, SDNE modified the objective function by adding another Laplacian eigenmaps term~\cite{belkin2002laplacian}:
\begin{equation}\label{eq:gae2}
        \min_{\mathbf{\Theta}} \mathcal{L}_2 + \alpha \sum \nolimits_{i,j=1}^N \mathbf{A}(i,j) \left\| \mathbf{h}_i - \mathbf{h}_j \right\|_2,
\end{equation}
i.e., two nodes also share similar latent representations if they are directly connected. The authors also modified the L2-reconstruction loss by using the adjacency matrix and assigning different weights to zero and non-zero elements:
\begin{equation}\label{eq:gae3}
        \mathcal{L}_2 = \sum \nolimits_{i=1}^N \left\| \left(\mathbf{A}\left(i,:\right) - \mathcal{G}\left( \mathbf{h}_i \right) \right) \odot \mathbf{b}_i \right\|_2,
\end{equation}
where $\mathbf{h}_i = \mathcal{F} \left( \mathbf{A}\left(i,: \right) \right)$, $b_{ij}=1$ if $\mathbf{A}(i,j)=0$; otherwise $b_{ij}=\beta > 1$, and $\beta$ is another hyper-parameter. The overall architecture of SDNE is shown in Figure~\ref{fig:SDNE}.

Motivated by another line of studies, a contemporary work DNGR~\cite{cao2016deep} replaced the transition matrix $\mathbf{P}$ in Eq.~\eqref{eq:gae1} with the positive pointwise mutual information (PPMI)~\cite{levy2014neural} matrix defined in Eq.~\eqref{eq:PPMI}. In this way, the raw features can be associated with some random walk probability of the graph~\cite{lovasz1993random}. However, constructing the input matrix has a time complexity of $O(N^2)$, which is not scalable to large-scale graphs.

GC-MC~\cite{berg2017graph} took a different approach by using the GCN proposed by Kipf and Welling~\cite{kipf2017semi} as the encoder:
\begin{equation}
   \mathbf{H} = GCN \left( \mathbf{F}^V,\mathbf{A} \right),
\end{equation}
and using a simple bilinear function as the decoder:
\begin{equation}
    \hat{\mathbf{A}}(i,j) = \mathbf{H}(i,:) \mathbf{\Theta}_{de} \mathbf{H}(j,:)^T,
\end{equation}
where $\mathbf{\Theta}_{de}$ are the decoder parameters. Using this approach, node features were naturally incorporated. For graphs without node features, a one-hot encoding of node IDs was utilized. The authors demonstrated the effectiveness of GC-MC on the recommendation problem on bipartite graphs.

Instead of reconstructing the adjacency matrix or its variations, DRNE~\cite{tu2018deep} proposed another modification that directly reconstructed the low-dimensional node vectors by aggregating neighborhood information using an LSTM. Specifically, DRNE adopted the following objective function:
\begin{equation}\label{eq:gnn4}
    \mathcal{L} = \sum \nolimits_{i=1}^N \left\| \mathbf{h}_i - \text{LSTM} \left( \{ \mathbf{h}_j | j \in \mathcal{N}(i)\} \right) \right\|.
\end{equation}
 Because an LSTM requires its inputs to be a sequence, the authors suggested ordering the node neighborhoods based on their degrees. They also adopted a neighborhood sampling technique for nodes with large degrees to prevent an overlong memory. The authors proved that such a method can preserve regular equivalence as well as many centrality measures of nodes, such as PageRank~\cite{page1999pagerank}.

Unlike the above works that map nodes into a low-dimensional vector, Graph2Gauss (G2G)~\cite{bojchevski2018deep} proposed encoding each node as a Gaussian distribution $\mathbf{h}_i = \mathcal{N}\left( \mathbf{M}(i,:),diag\left(\mathbf{\Sigma}(i,:)\right) \right)$ to capture the uncertainties of nodes. Specifically, the authors used a deep mapping from the node attributes to the means and variances of the Gaussian distribution as the encoder:
\begin{equation}
    \mathbf{M}(i,:) = \mathcal{F}_{\mathbf{M}}(\mathbf{F}^V(i,:)),\mathbf{\Sigma}(i,:) = \mathcal{F}_{\mathbf{\Sigma}}(\mathbf{F}^V(i,:)),
\end{equation}
where $\mathcal{F}_{\mathbf{M}}(\cdot)$ and $\mathcal{F}_{\mathbf{\Sigma}}(\cdot)$ are the parametric functions that need to be learned. Then, instead of using an explicit decoder function, they used pairwise constraints to learn the model:
\begin{equation}\label{eq:GAErank}
\begin{gathered}
    \text{KL}\left(\mathbf{h}_j || \mathbf{h}_i \right) < \text{KL}\left(\mathbf{h}_{j'} || \mathbf{h}_i \right) \\ \forall i, \forall j, \forall j' \; s.t. \; d(i,j) < d(i,j'),
\end{gathered}
\end{equation}
where $d(i,j)$ is the shortest distance from node $v_i$ to $v_j$ and $\text{KL}(q(\cdot) || p(\cdot))$ is the Kullback-Leibler (KL) divergence between $q(\cdot)$ and $p(\cdot)$~\cite{kullback1951information}. In other words, the constraints ensure that the KL-divergence between node representations has the same relative order as the graph distance.
However, because Eq.~\eqref{eq:GAErank} is hard to optimize, an energy-based loss~\cite{lecun2006tutorial} was adopted as a relaxation:
\begin{equation}
\mathcal{L} = \sum \nolimits_{(i,j,j')\in \mathcal{D}} \left( E_{ij}^2 + \exp^{-E_{ij'}}\right),
\end{equation}
where $\mathcal{D} = \left\{ (i,j,j')| d(i,j) < d(i,j') \right\}$ and $E_{ij} = \text{KL}(\mathbf{h}_j|| \mathbf{h}_i )$. The authors further proposed an unbiased sampling strategy to accelerate the training process.

\subsection{Variational Autoencoders}
Different from the aforementioned autoencoders, variational autoencoders (VAEs) are another type of deep learning method that combines dimensionality reduction with generative models. Its potential benefits include tolerating noise and learning smooth representations~\cite{kingma2014auto}. VAEs were first introduced to graph data in VGAE~\cite{kipf2016variational}, where the decoder was a simple linear product:
\begin{equation}
    p\left( \mathbf{A}|\mathbf{H}\right) = \prod \nolimits_{i,j=1}^N \sigma\left( \mathbf{h}_i \mathbf{h}_j^T \right),
\end{equation}
in which the node representation was assumed to follow a Gaussian distribution $q\left(\mathbf{h}_i |\mathbf{M},\mathbf{\Sigma} \right) = \mathcal{N}\left(\mathbf{h}_i | \mathbf{M}(i,:),diag\left( \mathbf{\Sigma}(i,:) \right) \right)$. For the encoder of the mean and variance matrices, the authors also adopted the GCN proposed by Kipf and Welling~\cite{kipf2017semi}:
\begin{equation}\label{eq:vgae1}
 \mathbf{M} = GCN_\mathbf{M}\left(\mathbf{F}^V,\mathbf{A} \right),
 \log \mathbf{\Sigma} = GCN_\mathbf{\Sigma}\left(\mathbf{F}^V,\mathbf{A} \right).
\end{equation}
Then, the model parameters were learned by minimizing the variational lower bound~\cite{kingma2014auto}:
\begin{equation}
\mathcal{L} = \mathbb{E}_{q\left( \mathbf{H} | \mathbf{F}^V,\mathbf{A} \right)} \left[ \log p\left( \mathbf{A}|\mathbf{H}\right)\right] - \text{KL}\left(q\left( \mathbf{H} | \mathbf{F}^V,\mathbf{A} \right) || p(\mathbf{H}) \right).
\end{equation}
However, because this approach required reconstructing the full graph, its time complexity is $O(N^2)$.

\begin{figure}
  \centering
  \includegraphics[width = 8cm]{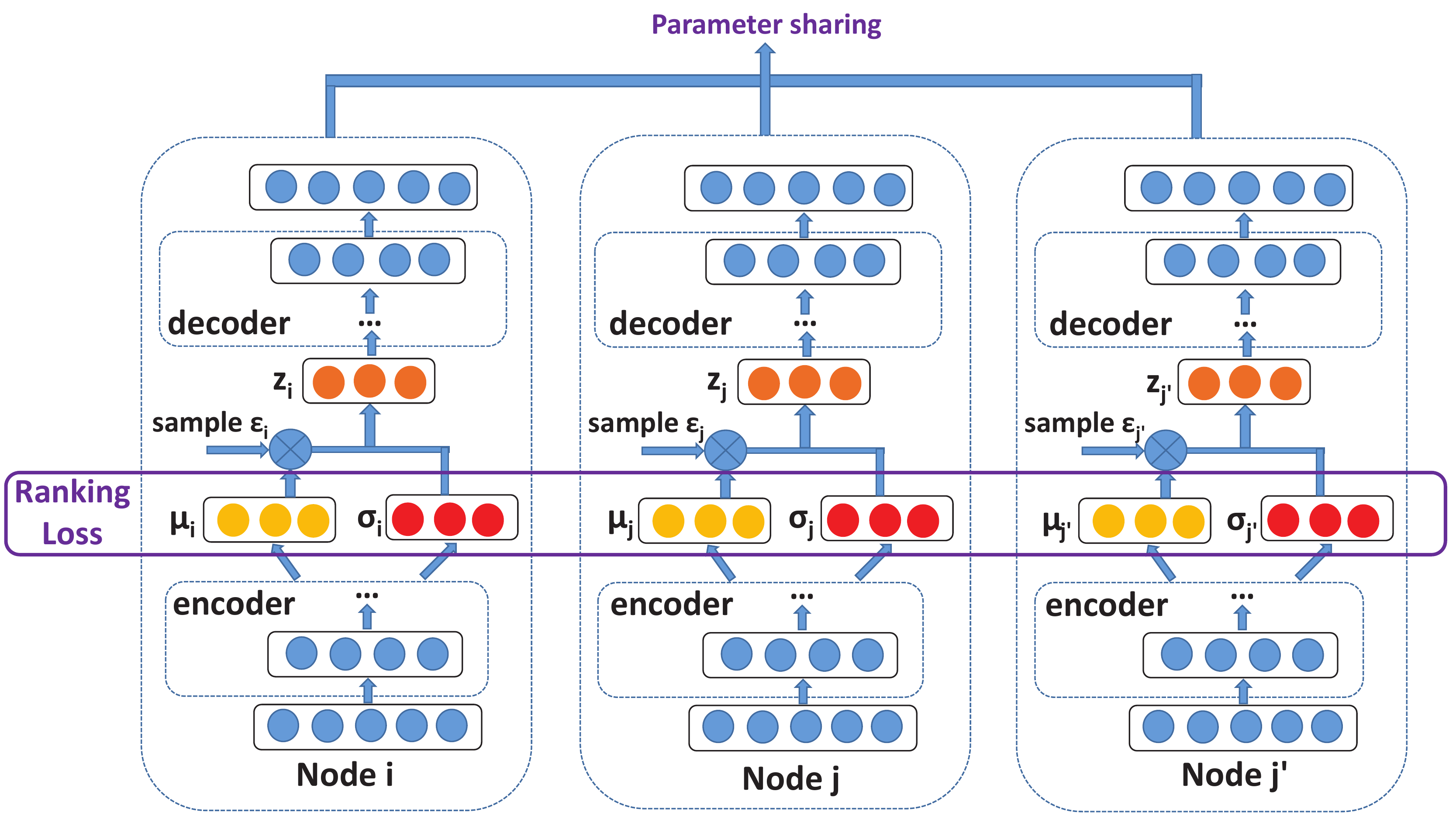}\\
  \caption{The framework of DVNE~\cite{zhu2018deep}. DVNE represents nodes as distributions using a VAE and adopts the Wasserstein distance to preserve the transitivity of the nodes similarities.}\label{fig:DVNE}
  \end{figure}

Motivated by SDNE and G2G, DVNE~\cite{zhu2018deep} proposed another VAE for graph data that also represented each node as a Gaussian distribution. Unlike the existing works that had adopted KL-divergence as the measurement, DVNE used the Wasserstein distance~\cite{vallender1974calculation} to preserve the transitivity of the nodes similarities. Similar to SDNE and G2G, DVNE also preserved both the first and second-order proximity in its objective function:
\begin{equation}
    \min_{\bf{\Theta}}  \sum \nolimits_{(i,j,j')\in \mathcal{D}} \left( E_{ij}^2 + \exp^{-E_{ij'}}\right) + \alpha \mathcal{L}_2,
\end{equation}
where $E_{ij}=W_2 \left( \mathbf{h}_j|| \mathbf{h}_i \right)$ is the $2^{nd}$ Wasserstein distance between two Gaussian distributions $\mathbf{h}_j$ and $\mathbf{h}_i$ and $\mathcal{D} = \left\{ (i,j,j')| j \in \mathcal{N}(i), j' \notin \mathcal{N}(i) \right\}$ is a set of triples corresponding to the ranking loss of the first-order proximity. The reconstruction loss was defined as follows:
\begin{equation}
    \mathcal{L}_2 = \inf \nolimits_{q(\mathbf{Z}|\mathbf{P})} \mathbb{E}_{p(\mathbf{P})} \mathbb{E}_{q(\mathbf{Z}|\mathbf{P})} \left\| \mathbf{P} \odot (\mathbf{P} - \mathcal{G}(\mathbf{Z}))\right\|_2^2,
\end{equation}
where $\mathbf{P}$ is the transition matrix and $\mathbf{Z}$ represents samples drawn from $\mathbf{H}$. The framework is shown in Figure~\ref{fig:DVNE}. Using this approach, the objective function can be minimized as in conventional VAEs using the reparameterization trick~\cite{kingma2014auto}.

\subsection{Improvements and Discussions}
Several improvements have also been proposed for GAEs.
\subsubsection{Adversarial Training}\label{sec:GAEgan}
\begin{figure*}
  \centering
  \includegraphics[width = 13.5cm]{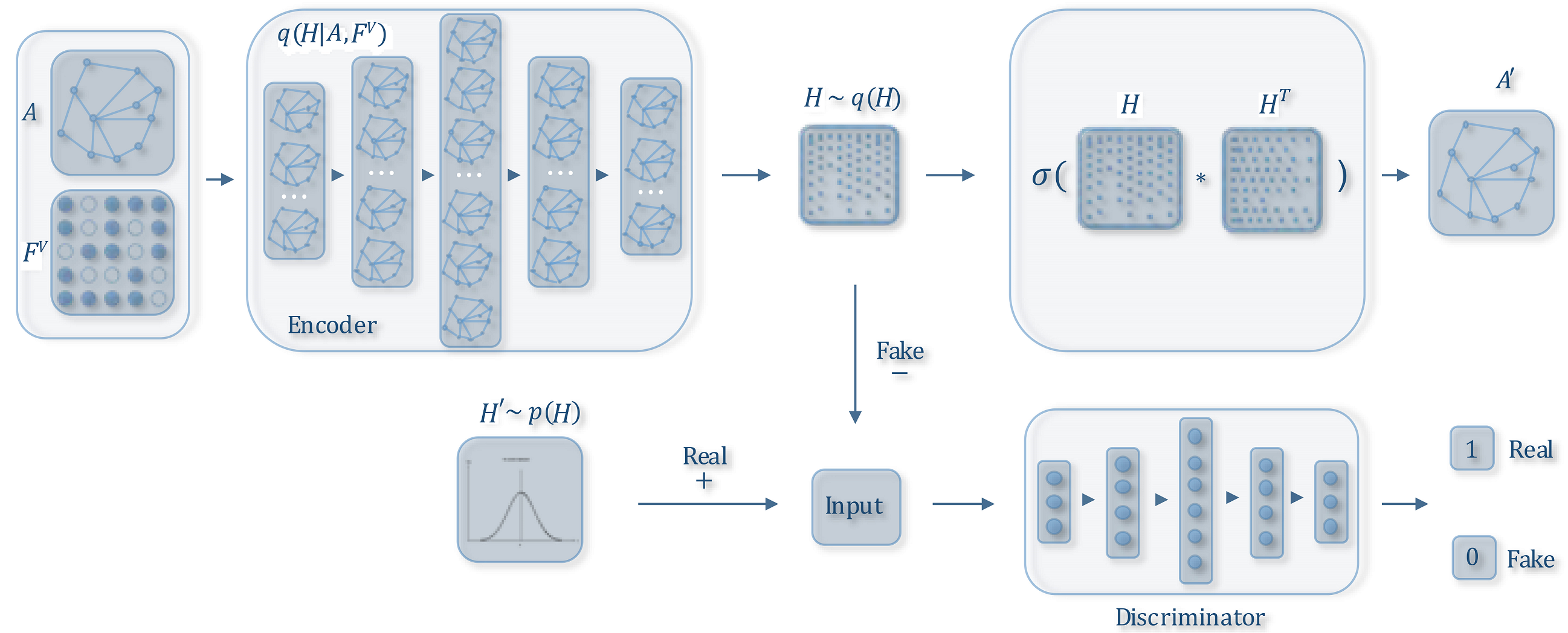}\\
  \caption{The framework of ARGA/ARVGA reprinted from~\cite{pan2018adversarially} with permission. This model incorporates the adversarial training scheme into GAEs.}\label{fig:ARGA}
  \end{figure*}

An adversarial training scheme\footnote{We will discuss more adversarial methods for graphs in Section~\ref{sec:adv}.} was incorporated into GAEs as an additional regularization term in ARGA~\cite{pan2018adversarially}. The overall architecture is shown in Figure~\ref{fig:ARGA}. Specifically, the encoder of GAEs was used as the generator while the discriminator aimed to distinguish whether a latent representation came from the generator or from a prior distribution. In this way, the autoencoder was forced to match the prior distribution as a regularization. The objective function was:
\begin{equation}
    \min_{\bf{\Theta}} \mathcal{L}_2 + \alpha \mathcal{L}_{GAN},
\end{equation}
where $\mathcal{L}_2$ is the reconstruction loss in GAEs and $\mathcal{L}_{GAN}$ is
\begin{equation}
    \min_{\mathcal{G}} \max_{\mathcal{D}} \mathbb{E}_{\mathbf{h} \sim p_{\mathbf{h}}} \left[ \log \mathcal{D}(\mathbf{h}) \right] + \mathbb{E}_{\mathbf{z} \sim \mathcal{G} ( \mathbf{F}^V,\mathbf{A} ) } \left[ \log\left( 1 - \mathcal{D} \left( \mathbf{z} \right) \right) \right],
\end{equation}
where $\mathcal{G}\left( \mathbf{F}^V,\mathbf{A}\right)$ is a generator that uses the graph convolutional encoder from Eq.~\eqref{eq:vgae1}, $\mathcal{D}(\cdot)$ is a discriminator based on the cross-entropy loss, and $p_{\mathbf{h}}$ is the prior distribution. The study adopted a simple Gaussian prior, and the experimental results demonstrated the effectiveness of the adversarial training scheme.

Concurrently, NetRA~\cite{yu2018learning} also proposed using a generative adversarial network (GAN)~\cite{goodfellow2014generative} to enhance the generalization ability of graph autoencoders. Specifically, the authors used the following objective function:
\begin{equation}
    \min_{\bf{\Theta}} \mathcal{L}_2 + \alpha_1 \mathcal{L}_{LE} + \alpha_2 \mathcal{L}_{GAN},
\end{equation}
where $\mathcal{L}_{LE}$ is the Laplacian eigenmaps objective function shown in Eq.~\eqref{eq:gae2}. In addition, the authors adopted an LSTM as the encoder to aggregate information from neighborhoods similar to Eq.~\eqref{eq:gnn4}. Instead of sampling only immediate neighbors and ordering the nodes using degrees as in DRNE~\cite{tu2018deep}, the authors used random walks to generate the input sequences. In contrast to ARGA, NetRA considered the representations in GAEs as the ground-truth and adopted random Gaussian noises followed by an MLP as the generator.

\subsubsection{Inductive Learning}
Similar to GCNs, GAEs can be applied to the inductive learning setting if node attributes are incorporated in the encoder. This can be achieved by using a GCN as the encoder, such as in GC-MC ~\cite{berg2017graph}, VGAE~\cite{kipf2016variational}, and VGAE~\cite{pan2018adversarially}, or by directly learning a mapping function from node features as in G2G~\cite{bojchevski2018deep}. Because the edge information is utilized only when learning the parameters, the model can also be applied to nodes unseen during training. These works also show that although GCNs and GAEs are based on different architectures, it is possible to use them jointly, which we believe is a promising future direction.

\subsubsection{Similarity Measures}
In GAEs, many similarity measures have been adopted, for example, L2-reconstruction loss, Laplacian eigenmaps, and the ranking loss for graph AEs, and KL divergence and Wasserstein distance for graph VAEs. Although these similarity measures are based on different motivations, how to choose an appropriate similarity measure for a given task and model architecture remains unstudied. More research is needed to understand the underlying differences between these metrics.

    \begin{table*}
    \centering
    \caption{The Main Characteristics of Graph Reinforcement Learning}\label{tab:GraphRL}
    \begin{tabular}{ | c | c | c | c | c |}
    \hline
    Method                       & Task & Actions                &  Rewards  & Time Complexity  \\ \hline
    GCPN~\cite{you2018graph}     & Graph generation & Link prediction        &  GAN + domain knowledge  & $O(MN)$    \\ \hline
    MolGAN~\cite{de2018molgan}   & Graph generation & Generate the entire graph &  GAN + domain knowledge  & $O(N^2)$    \\ \hline
    GTPN~\cite{do2019graph}      & Chemical reaction prediction & Predict node pairs and new bonding types & Prediction results & $O(N^2)$ \\ \hline
    GAM~\cite{lee2018graph}      & Graph classification & Predict graph labels and select the next node & Classification results & $O(d_{\text{avg}}sT)$ \\ \hline
    DeepPath~\cite{wenhan_emnlp2017} & Knowledge graph reasoning & Predict the next node of the reasoning path & Reasoning results + diversity & $O(d_{\text{avg}}sT + s^2T)$  \\ \hline
    MINERVA~\cite{minerva2018} & Knowledge graph reasoning & Predict the next node of the reasoning path & Reasoning results & $O(d_{\text{avg}}sT)$ \\ \hline
    \end{tabular}
    \end{table*}

\section{Graph Reinforcement Learning}\label{sec:RL}
One aspect of deep learning not yet discussed is reinforcement learning (RL), which has been shown to be effective in AI tasks such as playing games~\cite{silver2017mastering}. RL is known to be good at learning from feedbacks, especially when dealing with non-differentiable objectives and constraints. In this section, we review Graph RL methods. Their main characteristics are summarized in Table~\ref{tab:GraphRL}.

GCPN~\cite{you2018graph} utilized RL to generate goal-directed molecular graphs while considering non-differential objectives and constraints. Specifically, the graph generation is modeled as a Markov decision process of adding nodes and edges, and the generative model is regarded as an RL agent operating in the graph generation environment. By treating agent actions as link predictions, using domain-specific as well as adversarial rewards, and using GCNs to learn the node representations, GCPN can be trained in an end-to-end manner using a policy gradient~\cite{sutton2000policy}.

A concurrent work, MolGAN~\cite{de2018molgan}, adopted a similar idea of using RL for generating molecular graphs. However, rather than generating the graph through a sequence of actions, MolGAN proposed directly generating the full graph; this approach worked particularly well for small molecules.

GTPN~\cite{do2019graph} adopted RL to predict chemical reaction products. Specifically, the agent acted to select node pairs in the molecule graph and predicted their new bonding types, and rewards were given both immediately and at the end based on whether the predictions were correct. GTPN used a GCN to learn the node representations and an RNN to memorize the prediction sequence.

GAM~\cite{lee2018graph} applied RL to graph classification by using random walks. The authors modeled the generation of random walks as a partially observable Markov decision process (POMDP). The agent performed two actions: first, it predicted the label of the graph; then, it selected the next node in the random walk. The reward was determined simply by whether the agent correctly classified the graph, i.e.,
\begin{equation}
    \mathcal{J}(\theta) = \mathbb{E}_{P(S_{1:T};\theta)} \sum \nolimits_{t=1}^T r_t,
\end{equation}
where $r_t=1$ represents a correct prediction; otherwise, $r_t = -1$. $T$ is the total time steps and $S_t$ is the environment.

DeepPath~\cite{wenhan_emnlp2017} and MINERVA~\cite{minerva2018} both adopted RL for knowledge graph (KG) reasoning. Specifically, DeepPath targeted at pathfinding, i.e., find the most informative path between two target nodes, while MINERVA tackled question-answering tasks, i.e., find the correct answer node given a question node and a relation. In both methods, the RL agents need to predict the next node in the path at each step and output a reasoning path in the KG. Agents receive rewards if the paths reach the correct destinations. DeepPath also added a regularization term to encourage the path diversity.

    \begin{table*}
    \centering
    \caption{The Main Characteristics of Graph Adversarial Methods}\label{tab:GraphAdver}
    \begin{tabular}{ | c | c | c | c | c | }
    \hline
    Category                              & Method                                &~Adversarial Methods & Time Complexity & Node Features \\ \hline
    \multirow{8}{*}{\tabincell{c}{Adversarial\\Training}} & ARGA/ARVGA~\cite{pan2018adversarially}& Regularization for GAEs  & $O(N^2)$ & Yes  \\ \cline{2-5}
                                          & NetRA~\cite{yu2018learning}           & Regularization for GAEs  & $O(M)$ & No   \\ \cline{2-5}
                                          & GCPN~\cite{you2018graph}              & Rewards for Graph RL     & $O(MN)$  & Yes  \\ \cline{2-5}
                                          & MolGAN~\cite{de2018molgan}            & Rewards for Graph RL     & $O(N^2)$  & Yes  \\ \cline{2-5}
                                          & GraphGAN~\cite{wang2018graphgan}      & Generation of negative samples (i.e., node pairs) & $O(MN)$ & No \\ \cline{2-5}
                                          & ANE~\cite{dai2018adversarialne}       & Regularization for network embedding & $O(N)$ & No \\ \cline{2-5}
                                          & GraphSGAN~\cite{ding2018semi}         & Enhancing semi-supervised learning on graphs & $O(N^2)$ & Yes \\ \cline{2-5}
                                          & NetGAN~\cite{bojchevski2018netgan}    & Generation of graphs via random walks & $O(M)$ & No \\ \hline
    \multirow{3}{*}{\tabincell{c}{Adversarial\\Attack}}   & Nettack~\cite{zugner2018adversarial}  & Targeted attacks of graph structures and node attributes & $O(N d_0^2)$ & Yes \\  \cline{2-5}
                                          & Dai~\textit{et~al.}~\cite{dai2018adversarial}  & Targeted attacks of graph structures & $O(M)$ & No \\ \cline{2-5}
                                          & Zugner and Gunnemann~\cite{zugner2019adversarial} & Non-targeted attacks of graph structures & $O(N^2)$ & No \\
                                          \hline
    \end{tabular}
    \end{table*}

\section{Graph Adversarial Methods}\label{sec:adv}
Adversarial methods such as GANs~\cite{goodfellow2014generative} and adversarial attacks have drawn increasing attention in the machine learning community in recent years. In this section, we review how to apply adversarial methods to graphs. The main characteristics of graph adversarial methods are summarized in Table~\ref{tab:GraphAdver}.
\subsection{Adversarial Training}
The basic idea behind a GAN is to build two linked models: a discriminator and a generator. The goal of the generator is to ``fool'' the discriminator by generating fake data, while the discriminator aims to distinguish whether a sample comes from real data or is generated by the generator. Subsequently, both models benefit from each other by joint training using a minimax game. Adversarial training has been shown to be effective in generative models and enhancing the generalization ability of discriminative models. In Section~\ref{sec:GAEgan} and Section~\ref{sec:RL}, we reviewed how adversarial training schemes are used in GAEs and Graph RL, respectively. Here, we review several other adversarial training methods on graphs in detail.

GraphGAN~\cite{wang2018graphgan} proposed using a GAN to enhance graph embedding methods~\cite{cui2018survey} with the following objective function:
\begin{equation}\label{eq:GraphGAN}
\begin{aligned}
    \min_{\mathcal{G}} \max_{\mathcal{D}} \sum \nolimits_{i=1}^N & \left( \mathbb{E}_{v \sim p_{graph}(\cdot|v_i)} \left[ \log \mathcal{D}(v,v_i) \right] \right. \\
                          + & \left. \mathbb{E}_{v \sim \mathcal{G}(\cdot|v_i) } \left[ \log\left( 1 - \mathcal{D} \left(v,v_i \right) \right) \right] \right).
\end{aligned}
\end{equation}
The discriminator $\mathcal{D}(\cdot)$ and the generator $\mathcal{G}(\cdot)$ are as follows:
\begin{equation}
\begin{gathered}
    \mathcal{D}(v,v_i) = \sigma(\mathbf{d}_v \mathbf{d}_{v_i}^T), \mathcal{G}(v|v_i)= \frac{\exp(\mathbf{g}_v \mathbf{g}_{v_i}^T)}{\sum_{v^\prime \neq v_i}\exp(\mathbf{g}_{v^\prime} \mathbf{g}_{v_i}^T)},
\end{gathered}
\end{equation}
where $\mathbf{d}_v$ and $\mathbf{g}_v$ are the low-dimensional embedding vectors for node $v$ in the discriminator and the generator, respectively. Combining the above equations, the discriminator actually has two objectives: the node pairs in the original graph should possess large similarities, while the node pairs generated by the generator should possess small similarities. This architecture is similar to network embedding methods such as LINE~\cite{tang2015line}, except that negative node pairs are generated by the generator $\mathcal{G}(\cdot)$ instead of by random samplings. The authors showed that this method enhanced the inference abilities of the node embedding vectors.

Adversarial network embedding (ANE)~\cite{dai2018adversarialne} also adopted an adversarial training scheme to improve network embedding methods. Similar to ARGA~\cite{pan2018adversarially}, ANE used a GAN as an additional regularization term to existing network embedding methods such as DeepWalk~\cite{perozzi2014deepwalk} by imposing a prior distribution as the real data and regarding the embedding vectors as generated samples.

GraphSGAN~\cite{ding2018semi} used a GAN to enhance semi-supervised learning on graphs. Specifically, the authors observed that fake nodes should be generated in the density gaps between subgraphs to weaken the propagation effect across different clusters of the existing models. To achieve that goal, the authors designed a novel optimization objective with elaborate loss terms to ensure that the generator generated samples in the density gaps at equilibrium.

NetGAN~\cite{bojchevski2018netgan} adopted a GAN for graph generation tasks. Specifically, the authors regarded graph generation as a task to learn the distribution of biased random walks and adopted a GAN framework to generate and discriminate among random walks using an LSTM. The experiments showed that using random walks could also learn global network patterns.

\subsection{Adversarial Attacks}\label{sec:attack}
Adversarial attacks are another class of adversarial methods intended to deliberately ``fool'' the targeted methods by adding small perturbations to data. Studying adversarial attacks can deepen our understanding of the existing models and inspire more robust architectures. We review the graph-based adversarial attacks below.

Nettack~\cite{zugner2018adversarial} first proposed attacking node classification models such as GCNs by modifying graph structures and node attributes. Denoting the targeted node as $v_0$ and its true class as $c_{true}$, the targeted model as $\mathcal{F}(\mathbf{A},\mathbf{F}^V)$ and its loss function as $\mathcal{L}_{\mathcal{F}}(\mathbf{A},\mathbf{F}^V)$, the model adopted the following objective function:
\begin{equation}\label{eq:attack}
\begin{gathered}
        \argmax_{\left(\mathbf{A}^\prime,\mathbf{F}^{V\prime} \right) \in \mathcal{P} } \; \max_{c\neq c_{true}} \log \mathbf{Z}^*_{v_0,c} - \log \mathbf{Z}^*_{v_0,c_{true}} \\
        s.t. \; \mathbf{Z}^* = \mathcal{F}_{\theta^*}(\mathbf{A}^\prime,\mathbf{F}^{V\prime}),\theta^* = \argmin \nolimits_{\theta}\mathcal{L}_{\mathcal{F}}(\mathbf{A}^\prime,\mathbf{F}^{V\prime}),
\end{gathered}
\end{equation}
where $\mathbf{A}^\prime$ and $\mathbf{F}^{V\prime}$ are the modified adjacency matrix and node feature matrix, respectively, $\mathbf{Z}$ represents the classification probabilities predicted by $\mathcal{F}(\cdot)$, and $\mathcal{P}$ is the space determined by the attack constraints. Simply speaking, the optimization aims to find the best legitimate changes in graph structures and node attributes to cause $v_0$ to be misclassified. The $\theta^*$ indicates that the attack is causative, i.e., the attack occurs before training the targeted model. The authors proposed several constraints for the attacks. The most important constraint is that the attack should be ``unnoticeable'', i.e., it should make only small changes. Specifically, the authors proposed to preserve data characteristics such as node degree distributions and feature co-occurrences. The authors also proposed two attacking scenarios, direct attack (directly attacking $v_0$) and influence attack (only attacking other nodes), and several relaxations to make the optimization tractable.

Concurrently, Dai~\textit{et~al.}~\cite{dai2018adversarial} studied adversarial attacks for graphs with an objective function similar to Eq.~\eqref{eq:attack}; however, they focused on the case in which only graph structures were changed. Instead of assuming that the attacker possessed all the information, the authors considered several settings in which different amounts of information were available. The most effective strategy, RL-S2V, adopted structure2vec~\cite{dai2016discriminative} to learn the node and graph representations and used reinforcement learning to solve the optimization. The experimental results showed that the attacks were effective for both node and graph classification tasks.

The aforementioned two attacks are targeted, i.e., they are intended to cause misclassification of some targeted node $v_0$. Zugner and Gunnemann~\cite{zugner2019adversarial} were the first to study non-targeted attacks, which were intended to reduce the overall model performance. They treated the graph structure as hyper-parameters to be optimized and adopted meta-gradients in the optimization process, along with several techniques to approximate the meta-gradients.

\section{Discussions and Conclusion}
Thus far, we have reviewed the different graph-based deep learning architectures as well as their similarities and differences. Next, we briefly discuss their applications, implementations, and future directions before summarizing this paper.

\subsection{Applications}
In addition to standard graph inference tasks such as node or graph classification\footnote{A collection of methods for common tasks is listed in Appendix~\ref{sec:task}.}, graph-based deep learning methods have also been applied to a wide range of disciplines, including modeling social influence~\cite{qiu2018deepinf}, recommendation~\cite{monti2017geometric2,ying2018graph,berg2017graph,ma2019disentangle}, chemistry and biology~\cite{kearnes2016molecular,gilmer2017neural,duvenaud2015convolutional,you2018graph,de2018molgan}, physics~\cite{coley2017convolutional,xie2018crystal}, disease and drug prediction~\cite{ktena2017distance,zitnik2018modeling,parisot2017spectral}, gene expression~\cite{dutil2018towards}, natural language processing (NLP)~\cite{bastings2017graph,marcheggiani2017encoding}, computer vision~\cite{garcia2018few,Jain2016Structural,qi20173d,marino2017more,qi2018learning}, traffic forecasting~\cite{yu2018spatio,li2018diffusion}, program induction~\cite{allamanis2018learning}, solving graph-based NP problems~\cite{li2018combinatorial,prates2018learning}, and multi-agent AI systems~\cite{sukhbaatar2016learning,battaglia2016interaction,hoshen2017vain}.

A thorough review of these methods is beyond the scope of this paper due to the sheer diversity of these applications; however, we list several key inspirations. First, it is important to incorporate domain knowledge into the model when constructing a graph or choosing architectures. For example, building a graph based on the relative distance may be suitable for traffic forecasting problems, but may not work well for a weather prediction problem where the geographical location is also important. Second, a graph-based model can usually be built on top of other architectures rather than as a stand-alone model. For example, the computer vision community usually adopts CNNs for detecting objects and then uses graph-based deep learning as a reasoning module~\cite{santoro2017simple}. For NLP problems, GCNs can be adopted as syntactic constraints~\cite{bastings2017graph}. As a result, key key challenge is how to integrate different models. These applications also show that graph-based deep learning not only enables mining the rich value underlying the existing graph data but also helps to naturally model relational data as graphs, greatly widening the applicability of graph-based deep learning models.

\subsection{Implementations}
Recently, several open libraries have been made available for developing deep learning models on graphs. These libraries are listed in Table~\ref{tab:library}. We also collected a list of source code (mostly from their original authors) for the studies discussed in this paper. This repository is included in Appendix~\ref{sec:code}. These open implementations make it easy to learn, compare, and improve different methods. Some implementations also address the problem of distributed computing, which we do not discuss in this paper.

\subsection{Future Directions}
There are several ongoing or future research directions which are also worthy of discussion:
\begin{itemize}
\item \textbf{New models for unstudied graph structures}. Due to the extremely diverse structures of graph data, the existing methods are not suitable for all of them. For example, most methods focus on homogeneous graphs, while heterogeneous graphs are seldom studied, especially those containing different modalities such as those in~\cite{chang2015heterogeneous}. Signed networks, in which negative edges represent conflicts between nodes, also have unique structures, and they pose additional challenges to the existing methods~\cite{derr2018signed}. Hypergraphs, which represent complex relations between more than two objects~\cite{tu2018structural}, are also understudied. Thus, an important next step is to design specific deep learning models to handle these types of graphs.
\item \textbf{Compositionality of existing models}. As shown multiple times in this paper, many of the existing architectures can be integrated: for example, using a GCN as a layer in GAEs or Graph RL. In addition to designing new building blocks, how to systematically composite these architectures is an interesting future direction. In this process, how to incorporate interdisciplinary knowledge in a principled way rather than on a case-by-case basis is also an open problem. One recent work, graph networks~\cite{battaglia2018relational}, takes the first step and focuses on using a general framework of GNNs and GCNs for relational reasoning problems. AutoML may also be helpful by reducing the human burden of assembling different components and choosing hyper-parameters~\cite{tu2019autoNE}.
\item \textbf{Dynamic graphs}. Most of the existing methods focus on static graphs. However, many real graphs are dynamic in nature: their nodes, edges, and features can change over time. For example, in social networks, people may establish new social relations, remove old relations, and their features, such as hobbies and occupations, can change over time. New users may join the network and existing users may leave. How to model the evolving characteristics of dynamic graphs and support incremental updates to model parameters remain largely unaddressed. Some preliminary works have obtained encouraging results by using Graph RNNs~\cite{manessi2017dynamic,ma2018dynamic}.
\item \textbf{Interpretability and robustness}. Because graphs are often related to other risk-sensitive scenarios, the ability to interpret the results of deep learning models on graphs is critical in decision-making problems. For example, in medicine or disease-related problems, interpretability is essential in transforming computer experiments into applications for clinical use. However, interpretability for graph-based deep learning is even more challenging than are other black-box models because graph nodes and edges are often heavily interconnected. In addition, because many existing deep learning models on graphs are sensitive to adversarial attacks as shown in Section~\ref{sec:attack}, enhancing the robustness of the existing methods is another important issue. Some pioneering works regarding interpretability and robustness can be found in~\cite{ying2019gnn} and~\cite{zhu2019robust,jin2019power}, respectively.
\end{itemize}

\begin{table*}
\centering
\caption{Libraries of Deep Learning on Graphs}\label{tab:library}
\begin{tabular}{ | c | c | c | c|}
\hline
Name  & URL & Language/Framework & Key Characteristics \\ \hline
PyTorch Geometric~\cite{Fey2019Fast} & \url{https://github.com/rusty1s/pytorch_geometric} & PyTorch & \tabincell{c}{Improved efficiency, unified operations, \\ comprehensive existing methods} \\ \hline
Deep Graph Library~\cite{wang2019dgl} & \url{https://github.com/dmlc/dgl} & PyTorch & Improved efficiency, unified operations, scalability \\ \hline
AliGraph~\cite{zhu2019aligraph} & \url{https://github.com/alibaba/aligraph} & Unknown & Distributed environment, scalability, in-house algorithms \\ \hline
Euler~& \url{https://github.com/alibaba/euler} & C++/TensorFlow & Distributed environment, scalability \\ \hline
\end{tabular}
\end{table*}

\subsection{Summary}
The above survey shows that deep learning on graphs is a promising and fast-developing research field that both offers exciting opportunities and presents many challenges. Studying deep learning on graphs constitutes a critical building block in modeling relational data, and it is an important step towards a future with better machine learning and artificial intelligence techniques.

\section*{Acknowledgement}
The authors thank Jianfei Chen, Jie Chen, William L. Hamilton, Wenbing Huang, Thomas Kipf, Federico Monti, Shirui Pan, Petar Velickovic, Keyulu Xu, Rex Ying for allowing us to use their figures. This work was supported in part by National Program on Key Basic Research Project (No. 2015CB352300), National Key R\&D Program of China under Grand 2018AAA0102004, National Natural Science Foundation of China (No. U1936219, No. U1611461, No. 61772304), and Beijing Academy of Artificial Intelligence (BAAI). All opinions, findings, conclusions, and recommendations in this paper are those of the authors and do not necessarily reflect the views of the funding agencies.

\bibliographystyle{IEEEtran}
\bibliography{temp_cite}

\begin{IEEEbiography}[{\includegraphics[width=1in,height=1.25in,clip,keepaspectratio]{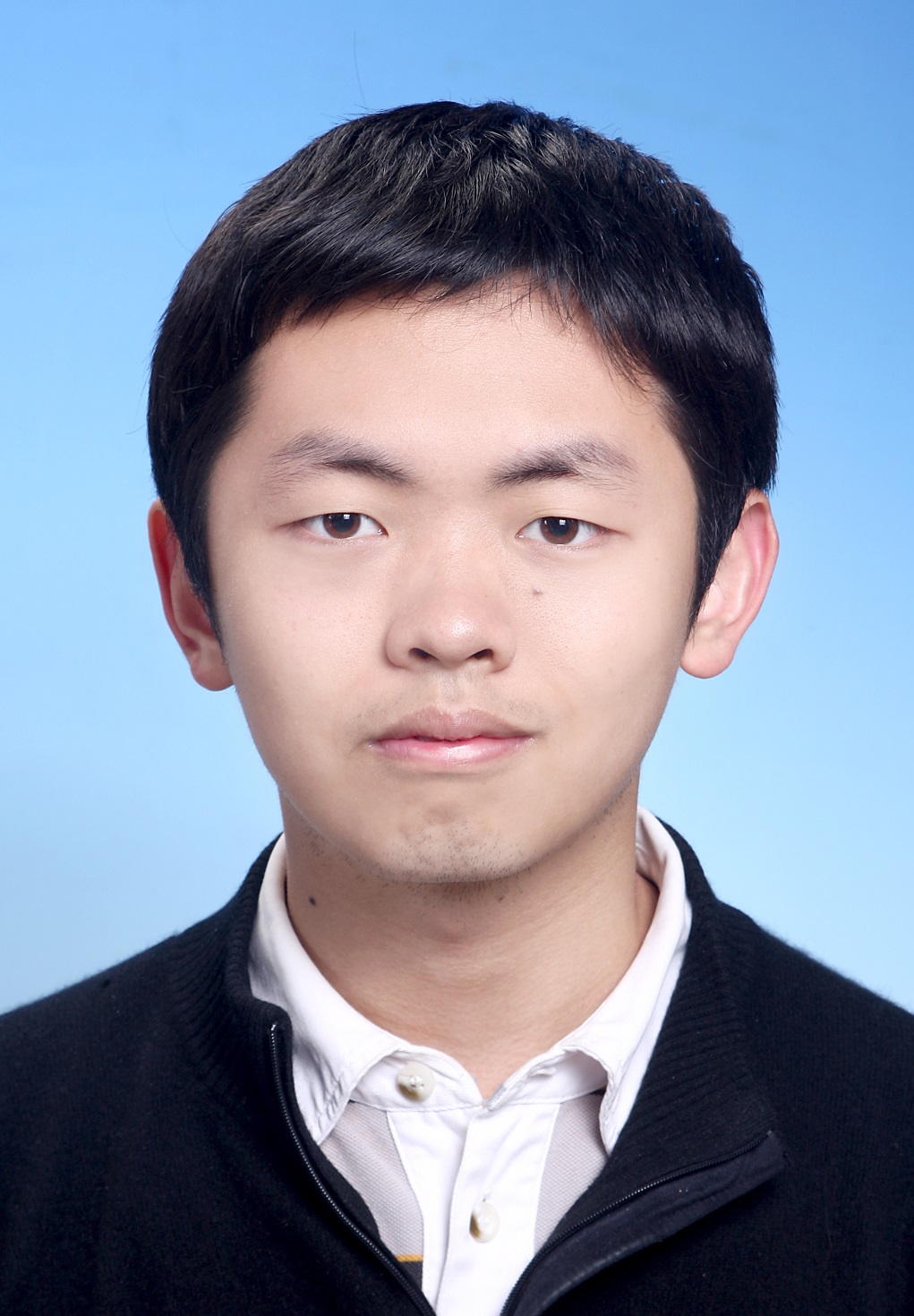}}]{Ziwei Zhang}
received his B.S. from the Department of Physics, Tsinghua University in 2016. He is currently pursuing a Ph.D. Degree in the Department of Computer Science and Technology at Tsinghua University. His research interests focus on network embedding and machine learning on graph data, especially in developing scalable algorithms for large-scale networks. He has published several papers in prestigious conferences and journals, including KDD, AAAI, IJCAI, and TKDE.
\end{IEEEbiography}

\begin{IEEEbiography}[{\includegraphics[width=1in,height=1.25in,clip,keepaspectratio]{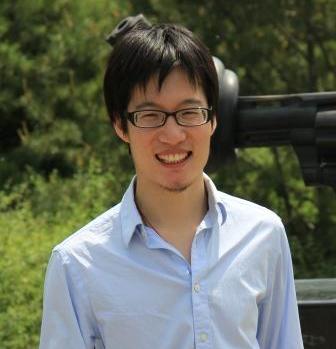}}]{Peng Cui}
received the Ph.D. degree from Tsinghua University in 2010. He is currently an associate professor with tenure at Tsinghua University. His research interests include network representation learning, human behavioral modeling, and social-sensed multimedia computing. He has published more than 100 papers in prestigious conferences and journals in data mining and multimedia. His recent research efforts have received the SIGKDD 2016 Best Paper Finalist, the ICDM 2015 Best Student Paper Award, the SIGKDD 2014 Best Paper Finalist, the IEEE ICME 2014 Best Paper Award, the ACM MM12 Grand Challenge Multimodal Award, and the MMM13 Best Paper Award. He is an associate editor of IEEE Transactions on Knowledge and Data Engineering, the IEEE Transactions on Big Data, the ACM Transactions on Multimedia Computing, Communications, and Applications, the Elsevier Journal on Neurocomputing, etc. He was the recipient of the ACM China Rising Star Award in 2015.
\end{IEEEbiography}

\begin{IEEEbiography}[{\includegraphics[width=1in,height=1.25in,clip,keepaspectratio]{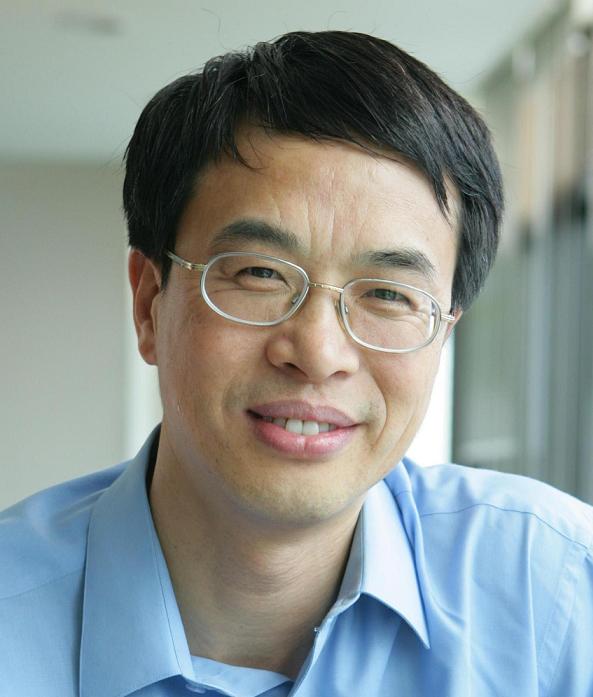}}]{Wenwu Zhu}
is currently a Professor and Deputy Head of the Computer Science Department of Tsinghua University and Vice Dean of National Research Center on Information Science and Technology. Prior to his current post, he was a Senior Researcher and Research Manager at Microsoft Research Asia. He was the Chief Scientist and Director at Intel Research China from 2004 to 2008. He worked at Bell Labs New Jersey as a Member of Technical Staff during 1996-1999. He received his Ph.D. degree from New York University in 1996.

He served as the Editor-in-Chief for the IEEE Transactions on Multimedia (T-MM) from January 1, 2017, to December 31, 2019. He has been serving as Vice EiC for IEEE Transactions on Circuits and Systems for Video Technology (TCSVT) and the chair of the steering committee for IEEE T-MM since January 1, 2020. His current research interests are in the areas of multimedia computing and networking, and big data. He has published over 400 papers in the referred journals and received nine Best Paper Awards including IEEE TCSVT in 2001 and 2019, and ACM Multimedia 2012. He is an IEEE Fellow, AAAS Fellow, SPIE Fellow and a member of the European Academy of Sciences (Academia Europaea).
\end{IEEEbiography}

\clearpage
\appendices
   \begin{table*}
    \centering
    \scriptsize
    \caption{A collection of published source code. O.A. = Original Authors}\label{tab:code}
    \begin{tabular}{ | c | c | l | c | c|}
    \hline
    Category & Method                                  & URL & O.A. & Language/Framework\\ \hline
    \multirow{4}{*}{Graph RNNs} & GGS-NNs~\cite{li2016gated} & \url{https://github.com/yujiali/ggnn} & Yes & Lua/Torch\\ \cline{2-5}
    & SSE~\cite{dai2018learning}                     & \url{https://github.com/Hanjun-Dai/steady_state_embedding} & Yes & C \\ \cline{2-5}
    & You~\textit{et~al.}~\cite{you2018graphrnn}              & \url{https://github.com/JiaxuanYou/graph-generation} & Yes & Python/PyTorch \\ \cline{2-5}
    & RMGCNN~\cite{monti2017geometric2}              & \url{https://github.com/fmonti/mgcnn}              & Yes & Python/TensorFlow \\ \hline
    \multirow{31}{*}{GCNs} & ChebNet \cite{defferrard2016convolutional} & \url{https://github.com/mdeff/cnn_graph}          & Yes & Python/TensorFlow\\ \cline{2-5}
    & Kipf\&Welling~\cite{kipf2017semi}           & \url{https://github.com/tkipf/gcn}                 & Yes & Python/TensorFlow\\ \cline{2-5}
    & CayletNet \cite{levie2017cayleynets}        & \url{https://github.com/amoliu/CayleyNet}          & Yes & Python/TensorFlow\\ \cline{2-5}
    & GWNN~\cite{xu2019graph}                     & \url{https://github.com/Eilene/GWNN}               & Yes & Python/TensorFlow\\ \cline{2-5}
    & Neural FPs~\cite{duvenaud2015convolutional} & \url{https://github.com/HIPS/neural-fingerprint}   & Yes & Python\\ \cline{2-5}
    & PATCHY-SAN~\cite{niepert2016learning}       & \url{https://github.com/seiya-kumada/patchy-san}   & No  & Python \\ \cline{2-5}
    & LGCN~\cite{gao2018large}                    & \url{https://github.com/divelab/lgcn/}             & Yes & Python/TensorFlow \\ \cline{2-5}
    & SortPooling~\cite{zhang2018end}             & \url{https://github.com/muhanzhang/DGCNN}          & Yes & Lua/Torch \\ \cline{2-5}
    & DCNN~\cite{atwood2016diffusion}             & \url{https://github.com/jcatw/dcnn}                & Yes & Python/Theano \\ \cline{2-5}
    & DGCN~\cite{zhuang2018dual}                  & \url{https://github.com/ZhuangCY/Coding-NN}        & Yes & Python/Theano \\ \cline{2-5}
    & MPNNs~\cite{gilmer2017neural}               & \url{https://github.com/brain-research/mpnn}       & Yes & Python/TensorFlow\\ \cline{2-5}
    & GraphSAGE~\cite{hamilton2017inductive}      & \url{https://github.com/williamleif/GraphSAGE}     & Yes & Python/TensorFlow\\ \cline{2-5}
    & GNs~\cite{battaglia2018relational}          & \url{https://github.com/deepmind/graph_nets}       & Yes & Python/TensorFlow\\ \cline{2-5}
    & DiffPool~\cite{ying2018hierarchical}        & \url{https://github.com/RexYing/graph-pooling}     & Yes & Python/PyTorch\\ \cline{2-5}
    & GAT~\cite{velickovic2018graph}              & \url{https://github.com/PetarV-/GAT}               & Yes & Python/TensorFlow \\ \cline{2-5}
    & GaAN~\cite{zhang2018gaan}                   & \url{https://github.com/jennyzhang0215/GaAN}       & Yes & Python/MXNet \\ \cline{2-5}
    & HAN~\cite{wang2019hetero}                   & \url{https://github.com/Jhy1993/HAN}               & Yes & Python/TensorFlow \\ \cline{2-5}
    & CLN~\cite{pham2017column}                   & \url{https://github.com/trangptm/Column_networks}  & Yes & Python/Keras\\ \cline{2-5}
    & PPNP~\cite{klicpera2019predict}             & \url{https://github.com/klicperajo/ppnp}           & Yes & Python/TensorFlow \\ \cline{2-5}
    & JK-Nets~\cite{xu2018representation}         & \url{https://github.com/mori97/JKNet-dgl}          & No  & Python/DGL \\ \cline{2-5}
    & ECC~\cite{simonovsky2017dynamic}            & \url{https://github.com/mys007/ecc}                & Yes & Python/PyTorch\\ \cline{2-5}
    & R-GCNs~\cite{schlichtkrull2018modeling}     & \url{https://github.com/tkipf/relational-gcn}      & Yes & Python/Keras \\ \cline{2-5}
    & LGNN~\cite{chen2019supervised}              & \url{https://github.com/joanbruna/GNN_community}   & Yes & Lua/Torch  \\ \cline{2-5}
    & StochasticGCN~\cite{chen2018stochastic}       & \url{https://github.com/thu-ml/stochastic_gcn}     & Yes & Python/TensorFlow \\ \cline{2-5}
    & FastGCN~\cite{chen2018fastgcn}              & \url{https://github.com/matenure/FastGCN}          & Yes & Python/TensorFlow\\ \cline{2-5}
    & Adapt~\cite{huang2018adaptive}              & \url{https://github.com/huangwb/AS-GCN}            & Yes & Python/TensorFlow \\ \cline{2-5}
    & Li~\textit{et~al.}~\cite{li2018deeper}               & \url{https://github.com/liqimai/gcn}               & Yes & Python/TensorFlow \\ \cline{2-5}
    & SGC~\cite{wu2019simplifying}                & \url{https://github.com/Tiiiger/SGC}               & Yes & Python/PyTorch \\ \cline{2-5}
    & GFNN~\cite{maehara2019revisiting}           & \url{https://github.com/gear/gfnn}                 & Yes & Python/PyTorch \\ \cline{2-5}
    & GIN~\cite{xu2019powerful}                   & \url{https://github.com/weihua916/powerful-gnns}   & Yes & Python/PyTorch \\ \cline{2-5}
    & DGI~\cite{velivckovic2019deep}              & \url{https://github.com/PetarV-/DGI}               & Yes & Python/PyTorch \\ \hline
    \multirow{10}{*}{GAEs} & SAE~\cite{tian2014learning} & \url{https://github.com/quinngroup/deep-representations-clustering} & No & Python/Keras \\ \cline{2-5}
    & SDNE~\cite{wang2016structural}                 & \url{https://github.com/suanrong/SDNE}             & Yes & Python/TensorFlow\\ \cline{2-5}
    & DNGR~\cite{cao2016deep}                        & \url{https://github.com/ShelsonCao/DNGR}           & Yes & Matlab\\ \cline{2-5}
    & GC-MC~\cite{berg2017graph}                     & \url{https://github.com/riannevdberg/gc-mc}        & Yes & Python/TensorFlow \\ \cline{2-5}
    & DRNE~\cite{tu2018deep}                         & \url{https://github.com/tadpole/DRNE}              & Yes & Python/TensorFlow \\ \cline{2-5}
    & G2G~\cite{bojchevski2018deep}                  & \url{https://github.com/abojchevski/graph2gauss}   & Yes & Python/TensorFlow \\ \cline{2-5}
    & VGAE~\cite{kipf2016variational}                & \url{https://github.com/tkipf/gae}                 & Yes & Python/TensorFlow \\ \cline{2-5}
    & DVNE~\cite{zhu2018deep}                        & \url{http://nrl.thumedialab.com}                   & Yes & Python/TensorFlow \\ \cline{2-5}
    & ARGA/ARVGA~\cite{pan2018adversarially}         & \url{https://github.com/Ruiqi-Hu/ARGA}             & Yes & Python/TensorFlow \\ \cline{2-5}
    & NetRA~\cite{yu2018learning}                    & \url{https://github.com/chengw07/NetRA}            & Yes & Python/PyTorch    \\ \hline
    \multirow{5}{*}{Graph RLs}  & GCPN~\cite{you2018graph} & \url{https://github.com/bowenliu16/rl_graph_generation} & Yes & Python/TensorFlow\\ \cline{2-5}
    & MolGAN~\cite{de2018molgan}                     & \url{https://github.com/nicola-decao/MolGAN}       & Yes & Python/TensorFlow \\ \cline{2-5}
    & GAM~\cite{lee2018graph}                        & \url{https://github.com/benedekrozemberczki/GAM}   & Yes & Python/Pytorhc \\ \cline{2-5}
    & DeepPath~\cite{wenhan_emnlp2017}               & \url{https://github.com/xwhan/DeepPath}            & Yes & Python/TensorFlow \\ \cline{2-5}
    & MINERVA~\cite{minerva2018}                     & \url{https://github.com/shehzaadzd/MINERVA}        & Yes & Python/TensorFlow \\ \hline
    \multirow{6}{*}{\tabincell{c}{Graph adversarial \\methods}} & GraphGAN~\cite{wang2018graphgan} & \url{https://github.com/hwwang55/GraphGAN} & Yes & Python/TensorFlow \\ \cline{2-5}
    & GraphSGAN~\cite{ding2018semi}                  & \url{https://github.com/dm-thu/GraphSGAN}          & Yes & Python/PyTorch\\ \cline{2-5}
    & NetGAN~\cite{bojchevski2018netgan}             & \url{https://github.com/danielzuegner/netgan}      & Yes & Python/TensorFlow\\ \cline{2-5}
    & Nettack~\cite{zugner2018adversarial}           & \url{https://github.com/danielzuegner/nettack}     & Yes & Python/TensorFlow\\ \cline{2-5}
    & Dai~\textit{et~al.}~\cite{dai2018adversarial}           & \url{https://github.com/Hanjun-Dai/graph_adversarial_attack} & Yes & Python/PyTorch \\ \cline{2-5}
    & Zugner\&Gunnemann~\cite{zugner2019adversarial} & \url{https://github.com/danielzuegner/gnn-meta-attack} & Yes & Python/TensorFlow\\
    \hline
    \multirow{21}{*}{Applications}
    & DeepInf~\cite{qiu2018deepinf}                  & \url{https://github.com/xptree/DeepInf}            & Yes & Python/PyTorch\\ \cline{2-5}
    & Ma~\textit{et~al.}~\cite{ma2019disentangle}    & \url{https://jianxinma.github.io/assets/disentangle-recsys-v1.zip} & Yes & Python/TensorFlow \\ \cline{2-5}
    & CGCNN~\cite{xie2018crystal}                    & \url{https://github.com/txie-93/cgcnn}             & Yes & Python/PyTorch\\ \cline{2-5}
    & Ktena~\textit{et~al.}~\cite{ktena2017distance} & \url{https://github.com/sk1712/gcn_metric_learning}& Yes & Python \\ \cline{2-5}
    & Decagon~\cite{zitnik2018modeling}              & \url{https://github.com/mims-harvard/decagon}      & Yes & Python/PyTorch\\ \cline{2-5}
    & Parisot~\textit{et~al.}~\cite{parisot2017spectral}& \url{https://github.com/parisots/population-gcn}& Yes & Python/TensorFlow \\ \cline{2-5}
    & Dutil~\textit{et~al.}~\cite{dutil2018towards}  & \url{https://github.com/mila-iqia/gene-graph-conv} & Yes & Python/PyTorch\\ \cline{2-5}
    & Bastings~\textit{et~al.}~\cite{bastings2017graph}& \url{https://github.com/bastings/neuralmonkey/tree/emnlp_gcn} & Yes & Python/TensorFlow \\ \cline{2-5}
    & Neural-dep-srl~\cite{marcheggiani2017encoding} & \url{https://github.com/diegma/neural-dep-srl}     & Yes & Python/Therano \\ \cline{2-5}
    & Garcia\&Bruna~\cite{garcia2018few}             & \url{https://github.com/vgsatorras/few-shot-gnn}   & Yes & Python/PyTorch\\ \cline{2-5}
    & S-RNN~\cite{Jain2016Structural}                & \url{https://github.com/asheshjain399/RNNexp}      & Yes & Python/Therano \\ \cline{2-5}
    & 3DGNN~\cite{qi20173d}                          & \url{https://github.com/xjqicuhk/3DGNN}            & Yes & Matlab/Caffe \\ \cline{2-5}
    & GPNN~\cite{qi2018learning}                     & \url{https://github.com/SiyuanQi/gpnn}             & Yes & Python/PyTorch \\ \cline{2-5}
    & STGCN~\cite{yu2018spatio}                      & \url{https://github.com/VeritasYin/STGCN_IJCAI-18} & Yes & Python/TensorFlow \\ \cline{2-5}
    & DCRNN~\cite{li2018diffusion}                   & \url{https://github.com/liyaguang/DCRNN}           & Yes & Python/TensorFlow \\ \cline{2-5}
    & Allamanis~\textit{et~al.}~\cite{allamanis2018learning} & \url{https://github.com/microsoft/tf-gnn-samples} & Yes & Python/TensorFlow \\ \cline{2-5}
    & Li~\textit{et~al.}~\cite{li2018combinatorial}  & \url{https://github.com/intel-isl/NPHard}          & Yes & Python/TensorFlow \\ \cline{2-5}
    & TSPGNN~\cite{prates2018learning}               & \url{https://github.com/machine-reasoning-ufrgs/TSP-GNN} & Yes & Python/TensorFlow \\ \cline{2-5}
    & CommNet~\cite{sukhbaatar2016learning}          & \url{https://github.com/facebookresearch/CommNet}  & Yes & Lua/Torch  \\ \cline{2-5}
    & Interaction network~\cite{battaglia2016interaction} & \url{https://github.com/jaesik817/Interaction-networks_tensorflow} & No & Python/TensorFlow \\ \cline{2-5}
    & Relation networks~\cite{santoro2017simple}     & \url{https://github.com/kimhc6028/relational-networks} & No & Python/PyTorch \\  \hline
    \multirow{6}{*}{Miscellaneous}
    & SGCN~\cite{derr2018signed}                     & \url{http://www.cse.msu.edu/~derrtyle/}            & Yes & Python/PyTorch\\ \cline{2-5}
    & DHNE~\cite{tu2018structural}                   & \url{https://github.com/tadpole/DHNE}              & Yes & Python/TensorFlow \\ \cline{2-5}
    & AutoNE~\cite{tu2019autoNE}                     & \url{https://github.com/tadpole/AutoNE}            & Yes & Python \\ \cline{2-5}
    & Gnn-explainer~\cite{ying2019gnn}               & \url{https://github.com/RexYing/gnn-model-explainer}&Yes & Python/PyTorch\\ \cline{2-5}
    & RGCN~\cite{zhu2019robust}                      & \url{https://github.com/thumanlab/nrlweb}          & Yes & Python/TensorFlow \\ \cline{2-5}
    & GNN-benchmark~\cite{shchur2018pitfalls}        & \url{https://github.com/shchur/gnn-benchmark}      & Yes & Python/TensorFlow \\ \hline
    \end{tabular}
    \end{table*}

\begin{table*}
\centering
\caption{A Table of Methods for Six Common Tasks}\label{tab:task}
\begin{tabular}{ | c | c | c | c | }
\cline{1-4}
Type                                                  & \multicolumn{2}{c|}{Task}                                                          & Methods \\ \cline{1-4}
\multirow{7}{*}{\tabincell{c}{Node-focused\\ Tasks}}  & \multicolumn{2}{c|}{Node Clustering}                                               &
                                                                                        \cite{tian2014learning,wang2019hetero,chang2015heterogeneous,wang2016structural,wang2018graphgan,
                                                                                        pan2018adversarially,levie2017cayleynets,cao2016deep} \\ \cline{2-4}
                                                                               & \multirow{2}{*}{Node Classification} & Transductive       & \tabincell{c}{\cite{henaff2015deep,scarselli2009graph,dai2018learning ,ma2018dynamic,defferrard2016convolutional,gao2018large,monti2017geometric,xu2019graph,zhuang2018dual,
                                       kipf2017semi,atwood2016diffusion,hamilton2017inductive,levie2017cayleynets,manessi2017dynamic} \\
                                       \cite{schlichtkrull2018modeling,klicpera2019predict,chen2019supervised,velickovic2018graph,wang2019hetero,pham2017column,
                                       chen2018fastgcn,chen2018stochastic,zhang2018gaan,li2018deeper,xu2018representation,huang2018adaptive,wu2019simplifying,maehara2019revisiting} \\
                                       \cite{zhu2018deep,bojchevski2018deep,velivckovic2019deep,tu2018deep,ding2018semi,wang2018graphgan,dai2018adversarialne,zhu2019robust,chang2015heterogeneous,
                                       wang2016structural,yu2018learning}} \\ \cline{3-4}
                                                                               &                                      & Inductive          &  \cite{velivckovic2019deep,gao2018large,velickovic2018graph,hamilton2017inductive,chen2018stochastic,zhang2018gaan,xu2018representation,
                                                                                dai2018learning,wu2019simplifying,bojchevski2018deep,maehara2019revisiting,chen2018fastgcn,huang2018adaptive}
                                                                               \\ \cline{2-4}
                                                                               & \multicolumn{2}{c|}{Network Reconstruction}               &    \cite{chang2015heterogeneous,wang2016structural,zhu2018deep,yu2018learning}
                                                                               \\ \cline{2-4}
                                                                               & \multicolumn{2}{c|}{Link Prediction}                      & \cite{kipf2016variational,berg2017graph,wang2016structural,bojchevski2018deep,wang2018graphgan,zhu2018deep,ying2018graph,monti2017geometric2,
                                                                                     schlichtkrull2018modeling,pan2018adversarially,levie2017cayleynets,ma2018dynamic,yu2018learning}
                                                                               \\ \cline{1-4}
\multirow{3}{*}{\tabincell{c}{Graph-focused\\ Tasks}} & \multicolumn{2}{c|}{Graph Classification}                                          &
                                                                               \cite{bruna2014spectral,henaff2015deep,manessi2017dynamic,levie2017cayleynets,zhang2018end,ying2018hierarchical,kearnes2016molecular,lee2018graph,dai2016discriminative,niepert2016learning,atwood2016diffusion,
                                                                               xu2019powerful,wu2019simplifying,simonovsky2017dynamic,defferrard2016convolutional,
                                                                               scarselli2009graph,monti2017geometric}       \\ \cline{2-4}
                                                                               & \multirow{2}{*}{Graph Generation}    & Structure-only     &
                                                                               \cite{bojchevski2018netgan,you2018graphrnn}      \\ \cline{3-4}
                                                                               &                                      & Structure+features &
                                                                               \cite{de2018molgan,you2018graph,li2018learning}  \\ \cline{1-4}
\end{tabular}
\end{table*}

\section{Source Codes}\label{sec:code}
Table~\ref{tab:code} shows a collection and summary of the source code we collected for the papers discussed in this manuscript. In addition to method names and links, the table also lists the programming language used and the frameworks adopted as well as whether the code was published by the original authors of the paper.

\section{Applicability for Common Tasks}\label{sec:task}
Table~\ref{tab:task} summarizes the applicability of different models for six common graph tasks, including node clustering, node classification, network reconstruction, link prediction, graph classification, and graph generation. Note that these results are based on whether the experiments were reported in the original papers.

\section{Node Classification Results on Benchmark Datasets}
As shown in Appendix~\ref{sec:task}, node classification is the most common task for graph-based deep learning models. Here, we report the results of different methods on five node classification benchmark datasets\footnote{These five benchmark datasets are publicly available at \url{https://github.com/tkipf/gcn} or \url{http://snap.stanford.edu/graphsage/}.}:
\begin{itemize}
\item Cora, Citeseer, PubMed~\cite{sen2008collective}: These are citation graphs with nodes representing papers, edges representing citations between papers, and papers associated with bag-of-words features and ground-truth topics as labels.
\item Reddit~\cite{hamilton2017inductive}: Reddit is an online discussion forum in which nodes represent posts and two nodes are connected when they are commented by the same user, and each post contains a low-dimensional word vector as features and a label indicating the Reddit community in which it was posted.
\item PPI~\cite{hamilton2017inductive}: PPI is a collection of protein-protein interaction graphs for different human tissues. It includes features that represent biological signatures and labels that represent the roles of proteins.
\end{itemize}
Cora, Citeseer, and Pubmed each include one graph, and the same graph structure is used for both training and testing, thus the tasks are considered transductive. In Reddit and PPI, because the training and testing graphs are different, these two datasets are considered to be inductive node classification benchmarks.

In Table~\ref{tab:benchmark}, we report the results of different models on these benchmark datasets. The results were extracted from their original papers when a fixed dataset split was adopted. The table shows that many state-of-the-art methods achieve roughly comparable performance on these benchmarks, with differences smaller than one percent. Shchur~\textit{et~al.}~\cite{shchur2018pitfalls} also found that a fixed dataset split can easily result in spurious comparisons. As a result, although these benchmarks have been widely adopted to compare different models, more comprehensive evaluation setups are critically needed.

\begin{table*}
    \centering
    \caption{Statistics of the benchmark datasets and the node classification results of different methods when a fixed dataset split is adopted. A hyphen ('-') indicates that the result is unavailable in the paper.}
    \label{tab:benchmark}
    \begin{tabular}{| c | c | c | c | c | c |} \hline
             & Cora     & Citeseer & Pubmed    & Reddit    & PPI     \\ \hline
    Type     & Citation & Citation & Citation  & Social    & Biology \\
    Nodes    & 2,708    & 3,327    & 19,717    & 232,965   & 56,944 (24 graphs)\\
    Edges    & 5,429    &  4,732   & 44,338    & 11,606,919& 818,716 \\
    Classes  & 7        & 6        & 3         & 41        & 121     \\
    Features & 1,433    & 3,703    & 500       & 602       & 50      \\
    Task     & Transductive & Transductive & Transductive & Inductive & Inductive \\ \hline
    Bruna~\textit{et~al.}~\cite{bruna2014spectral}\footnotemark & 73.3 & 58.9 & 73.9 & - & -\\
    ChebNet~\cite{defferrard2016convolutional}\footnotemark & 81.2 & 69.8 & 74.4 & - & - \\
    GCN~\cite{kipf2017semi}                    & 81.5     & 70.3     & 79.0     & -        & -        \\
    CayleyNets~\cite{levie2017cayleynets}      & 81.9$\pm$0.7 & -        & -    & -        & -        \\
    GWNN~\cite{xu2019graph}                    & 82.8     & 71.7     & 79.1     & -        & -        \\
    LGCN~\cite{gao2018large}                   & 83.3$\pm$0.5 & 73.0$\pm$0.6 & 79.5$\pm$0.2 & -        & 77.2$\pm$0.2 \\
    DGCN~\cite{zhuang2018dual}                 & 83.5     & 72.6     & 80.0     & -        & -        \\
    GraphSAGE~\cite{hamilton2017inductive}     & -        & -        & -        & 95.4     & 61.2     \\
    MoNet~\cite{monti2017geometric}            & 81.7$\pm$0.5 & -        & 78.8$\pm$0.4 & -        & -        \\
    GAT~\cite{velickovic2018graph}             & 83.0$\pm$0.7 & 72.5$\pm$0.7 & 79.0$\pm$0.3 & -        & 97.3$\pm$0.2 \\
    GaAN~\cite{zhang2018gaan}                  & -        & -        & -        & 96.4$\pm$0.0 & 98.7$\pm$0.0 \\
    JK-Nets~\cite{xu2018representation}        & -        & -        & -        & 96.5     & 97.6$\pm$0.7 \\
    StochasticGCN~\cite{chen2018stochastic}    & 82.0$\pm$0.8 & 70.9$\pm$0.2 & 79.0$\pm$0.4 & 96.3$\pm$0.0 & 97.9$\pm$0.0 \\
    FastGCN~\cite{chen2018fastgcn}             & 72.3     & -        & 72.1     & 93.7     & -        \\
    Adapt~\cite{huang2018adaptive}             & -        & -        & -        & 96.3$\pm$0.3 & -        \\
    SGC~\cite{wu2019simplifying}               & 81.0$\pm$0.0 & 71.9$\pm$0.1 & 78.9$\pm$0.0 & 94.9     & -        \\
    DGI~\cite{velivckovic2019deep}             & 82.3$\pm$0.6 & 71.8$\pm$0.7 & 76.8$\pm$0.6 & 94.0$\pm$0.1 & 63.8$\pm$0.2 \\
    SSE~\cite{dai2018learning}                 & -        & -        & -        & -        & 83.6     \\
    GraphSGAN~\cite{ding2018semi}              & 83.0$\pm$1.3 & 73.1$\pm$1.8 & -        & -        & -        \\
    RGCN~\cite{zhu2019robust}                  & 82.8$\pm$0.6 & 71.2$\pm$0.5 & 79.1$\pm$0.3 & -        & -        \\\hline
    \end{tabular}
    \end{table*}
\begin{figure}
  \centering
  \includegraphics[width=8.5cm]{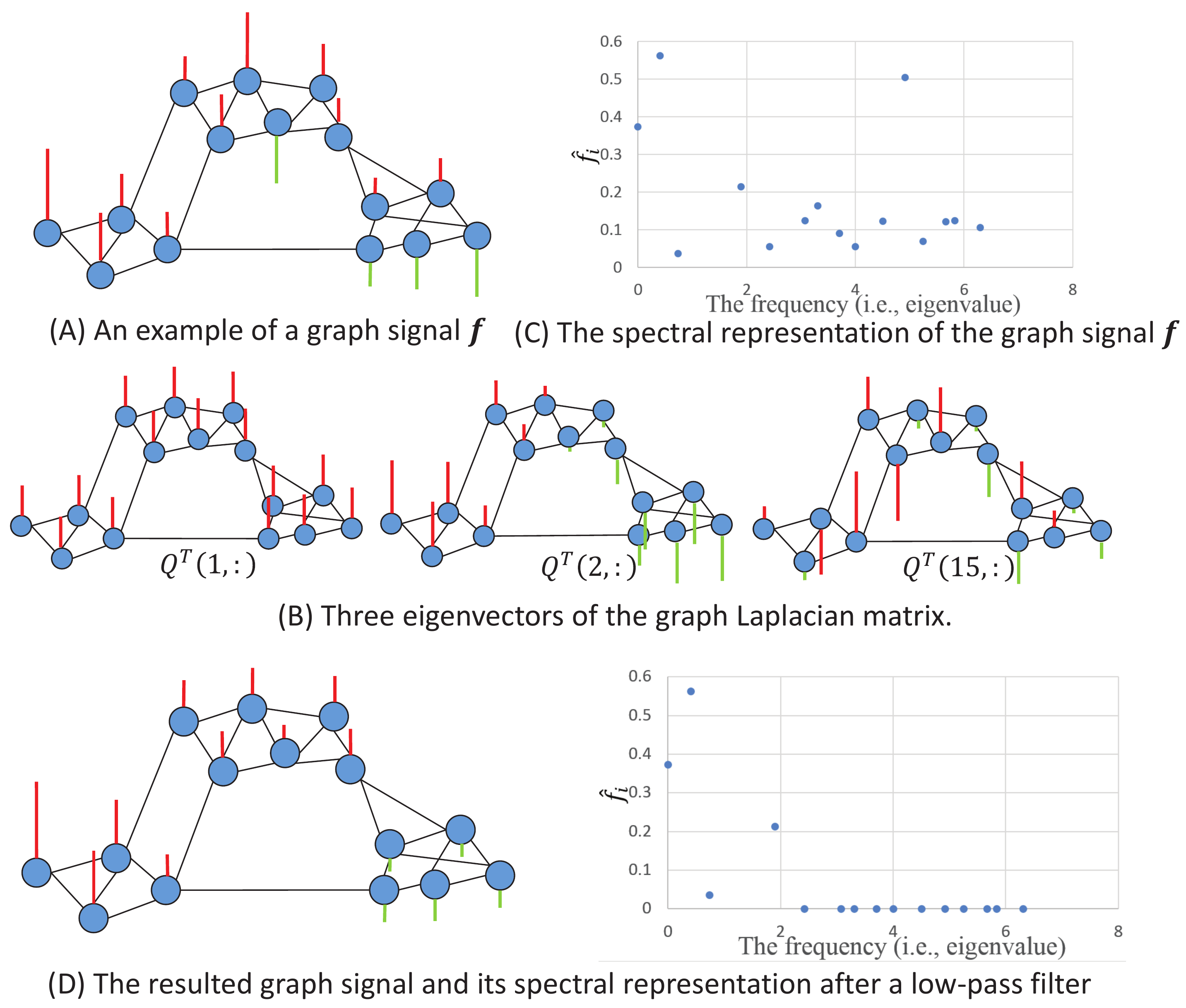}\\
  \caption{An example of graph signals and its spectral representations transformed using the eigenvectors of the graph Laplacian matrix. The upward-pointing red lines represent positive values and the downward-pointing green lines represent negative values. These images were adapted from \cite{shuman2013emerging}.}\label{fig:signal}
\end{figure}

\section{An Example of Graph Signals}\label{sec:signal}
To help understanding GCNs, we provide an example of graph signals and refer readers to \cite{shuman2013emerging,bronstein2017geometric,ortega2018graph} for more comprehensive surveys.

Given a graph $G=(V,E)$, a graph signal $\mathbf{f}$ corresponds to a collection of numbers: one number for each node in the graph. For undirected graphs, we usually assume that the signal takes real values, i.e., $\mathbf{f} \in \mathbb{R}^{N}$, where $N$ is the number of nodes. Any node feature satisfying the above requirement can be regarded as a graph signal, with
an example shown in Fig~\ref{fig:signal}~(A). Both the signal values and the underlying graph structure are important in processing and analyzing graph signals. For example, we can transform a graph signal into the spectral domain using the eigenvectors of the graph Laplacian matrix:
\begin{equation}
    \hat{\mathbf{f}} = \mathbf{Q}^T \mathbf{f}
\end{equation}
or equivalently
\begin{equation}
    \hat{\mathbf{f}}_i = \mathbf{Q}^T(i,:) \mathbf{f}.
\end{equation}
Because the eigenvectors $\mathbf{Q}^T$ are sorted in ascending order based on their corresponding eigenvalues, it has been shown~\cite{shuman2013emerging} that they form a basis for graph signals based on different ``smoothness''. Specifically, eigenvectors corresponding to small eigenvalues represent smooth signals and low frequencies, while eigenvectors corresponding to large eigenvalues represent non-smooth signals and high frequencies, as shown in Fig~\ref{fig:signal}~(B). Note that the smoothness is measured with respect to the graph structure, i.e., whether the signals oscillate across edges in the graph. As a result, $\hat{\mathbf{f}}$ provides a spectral representation of the signal $\mathbf{f}$ as shown in Fig~\ref{fig:signal}~(C). This is similar to the Fourier transform in Euclidean spaces. Using $\hat{\mathbf{f}}$, we can design various signal processing operations. For example, if we apply a low-pass filter, the resulted signal will be more smooth, as shown in Fig~\ref{fig:signal}~(D) (in this example, we set the frequency threshold as 2, i.e., only keeping the lowest 4 frequencies).

\newpage
\section{Time Complexity}
\footnotetext{The results were reported in GWNN~\cite{xu2019graph}.}
\footnotetext{The results were reported in Kipf and Welling~\cite{kipf2016variational}.}
In this section, we explain how we obtained the time complexity in all the tables. Specifically, we mainly focus on the time complexity with respect to the graph size, e.g., the number of nodes $N$ and the number of edges $M$, and omit other factors, e.g., the number of hidden dimensions $f_l$ or the number of iterations, since the latter terms are usually set as small constants and are less dominant.
Note that we focus on the theoretical results, while the exact efficiency of one algorithm also depends heavily on its implementations and techniques to reduce the constants in the time complexity.

\begin{itemize}
\item GNN~\cite{scarselli2009graph}: $O(MI_f)$, where $I_f$ is the number of iterations for Eq.~\eqref{eq:gnn1} to reach stable points, as shown in the paper.
\item GGS-NNs~\cite{li2016gated}: $O(MT)$, where $T$ is a preset maximum pseudo time since the method utilizes all the edges in each updating.
\item SSE~\cite{dai2018learning}: $O(d_{\text{avg}}S)$, where $d_{\text{avg}}$ is the average degree and $S$ is the total number of samples, as shown in the paper.
\item You~\textit{et~al.}~\cite{you2018graphrnn}: $O(N^2)$, as shown in the paper.
\item DGNN~\cite{ma2018dynamic}: $O(Md_{\text{avg}})$, where $d_{\text{avg}}$ is the average degree since the effect of the one-step propagation of each edge is considered.
\item RMGCNN~\cite{monti2017geometric2}: $O(MN)$ or $O(M)$, depending on whether an approximation technique is adopted, as shown in the paper.
\item Dynamic GCN~\cite{manessi2017dynamic}: $O(Mt)$, where $t$ denotes the number of time slices since the model runs one GCN at each time slice.
\item Bruna~\textit{et~al.}~\cite{bruna2014spectral} and Henaff~\textit{et~al.}~\cite{henaff2015deep}: $O(N^3)$, due to the time complexity of the eigendecomposition.
\item ChebNet~\cite{defferrard2016convolutional}, Kipf and Welling~\cite{kipf2017semi}, CayletNet~\cite{levie2017cayleynets}, GWNN~\cite{xu2019graph}, and Neural FPs~\cite{duvenaud2015convolutional}: $O(M)$, as shown in the corresponding papers.
\item PATCHY-SAN~\cite{niepert2016learning}: $O(M\log N)$, assuming the method adopts WL to label nodes, as shown in the paper.
\item LGCN~\cite{gao2018large}: $O(M)$ since all the neighbors of each node are sorted in the method.
\item SortPooling~\cite{zhang2018end}: $O(M)$, due to the time complexity of adopted graph convolution layers.
\item DCNN~\cite{atwood2016diffusion}: $O(N^2)$, as reported in~\cite{kipf2017semi}.
\item DGCN~\cite{zhuang2018dual}: $O(N^2)$ since the PPMI matrix is not sparse.
\item MPNNs~\cite{gilmer2017neural}: $O(M)$, as shown in the paper.
\item GraphSAGE~\cite{hamilton2017inductive}: $O(Ns^L)$, where $s$ is the size of the sampled neighborhoods and $L$ is the number of layers, as shown in the paper.
\item MoNet~\cite{monti2017geometric}: $O(M)$ since only the existing node pairs are involved in the calculation.
\item GNs~\cite{battaglia2018relational}: $O(M)$ since only the existing node pairs are involved in the calculation.
\item Kearnes~\textit{et~al.}~\cite{kearnes2016molecular}: $O(M)$, since only the existing node pairs are used in the calculation.
\item DiffPool~\cite{ying2018hierarchical}: $O(N^2)$ since the coarsened graph is not sparse.
\item GAT~\cite{velickovic2018graph}: $O(M)$, as shown in the paper.
\item GaAN~\cite{zhang2018gaan}: $O(Ns^L)$, where $s$ is a preset maximum neighborhood length and $L$ is the number of layers, as shown in the paper.
\item HAN~\cite{wang2019hetero}: $O(M_\phi)$, the number of meta-path-based node pairs, as shown in the paper.
\item CLN~\cite{pham2017column}: $O(M)$ since only the existing node pairs are involved in the calculation.
\item PPNP~\cite{klicpera2019predict}: $O(M)$, as shown in the paper.
\item JK-Nets~\cite{xu2018representation}: $O(M)$, due to the time complexity in adopted graph convolutional layers.
\item ECC~\cite{simonovsky2017dynamic}: $O(M)$, as shown in the paper.
\item R-GCNs~\cite{schlichtkrull2018modeling}: $O(M)$ since the edges of different types sum up to the total number of edges of the graph.
\item LGNN~\cite{chen2019supervised}: $O(M)$, as shown in the paper.
\item PinSage~\cite{ying2018graph}: $O(Ns^L)$, where $s$ is the size of the sampled neighborhoods and $L$ is the number of layers since a sampling strategy similar to that of GraphSAGE~\cite{hamilton2017inductive} is adopted.
\item StochasticGCN~\cite{chen2018stochastic}: $O(Ns^L)$, as shown in the paper.
\item FastGCN~\cite{chen2018fastgcn} and Adapt~\cite{huang2018adaptive}: $O(NsL)$ since the samples are drawn in each layer instead of in the neighborhoods, as shown in the paper.
\item Li~\textit{et~al.}~\cite{li2018deeper}: $O(M)$, due to the time complexity in adopted graph convolutional layers.
\item SGC~\cite{wu2019simplifying}: $O(M)$ since the calculation is the same as Kipf and Welling~\cite{kipf2017semi} by not adopting nonlinear activations.
\item GFNN~\cite{maehara2019revisiting}: $O(M)$ since the calculation is the same as SGC~\cite{wu2019simplifying} by adding an extra MLP layer.
\item GIN~\cite{xu2019powerful}: $O(M)$, due to the time complexity in adopted graph convolutional layers.
\item DGI~\cite{velivckovic2019deep}: $O(M)$, due to the time complexity in adopted graph convolutional layers.
\item SAE~\cite{tian2014learning} and SDNE~\cite{wang2016structural}: $O(M)$, as shown in the corresponding papers.
\item DNGR~\cite{cao2016deep}: $O(N^2)$, due to the time complexity of calculating the PPMI matrix.
\item GC-MC~\cite{berg2017graph}: $O(M)$ since the encoder adopts the GCN proposed by Kipf and Welling~\cite{kipf2017semi} and only the non-zero elements of the graph are considered in the decoder.
\item DRNE~\cite{tu2018deep}: $O(Ns)$, where $s$ is a preset maximum neighborhood length, as shown in the paper.
\item G2G~\cite{bojchevski2018deep}: $O(M)$, due to the definition of the ranking loss.
\item VGAE~\cite{kipf2016variational}: $O(N^2)$, due to the reconstruction of all the node pairs.
\item DVNE~\cite{zhu2018deep}: Though the original paper reported to have a time complexity of $O(M d_{\text{avg}})$ where $d_{\text{avg}}$ is the average degree, we have confirmed that it can be easily improved to $O(M)$ through personal communications with the authors.
\item ARGA/ARVGA~\cite{pan2018adversarially}: $O(N^2)$, due to the reconstruction of all the node pairs.
\item NetRA~\cite{yu2018learning}: $O(M)$, as shown in the paper.
\item GCPN~\cite{you2018graph}: $O(MN)$ since the embedding of all the nodes are used when generating each edge.
\item MolGAN~\cite{de2018molgan} and GTPN~\cite{do2019graph}: $O(N^2)$ since the scores for all the node pairs have to be calculated.
\item GAM~\cite{lee2018graph}: $O(d_{\text{avg}}sT)$, where $d_{\text{avg}}$ is the average degree, $s$ is the number of sampled random walks, and $T$ is the walk length, as shown in the paper.
\item DeepPath~\cite{wenhan_emnlp2017}: $O(d_{\text{avg}}sT + s^2T)$, where $d_{\text{avg}}$ is the average degree, $s$ is the number of sampled paths, and $T$ is the path length. The former term corresponds to finding paths and the latter term results from the diversity constraint.
\item MINERVA~\cite{minerva2018}: $O(d_{\text{avg}}sT)$, where $d_{\text{avg}}$ is the average degree, $s$ is the number of sampled paths, and $T$ is the path length, similar to the pathfinding method in DeepPath~\cite{wenhan_emnlp2017}.
\item GraphGAN~\cite{wang2018graphgan}: $O(MN)$, as shown in the paper.
\item ANE~\cite{dai2018adversarialne}: $O(N)$, which is the extra time complexity introduced by the model in the generator and the discriminator.
\item GraphSGAN~\cite{ding2018semi}: $O(N^2)$, due to the time complexity in the objective function.
\item NetGAN~\cite{bojchevski2018netgan}: $O(M)$, as shown in the paper.
\item Nettack~\cite{zugner2018adversarial}: $O(N d_0^2)$, where $d_0$ is the degree of the targeted node, as shown in the paper.
\item Dai~\textit{et~al.}~\cite{dai2018adversarial}: $O(M)$, which is the time complexity of the most effective strategy RL-S2V, as shown in the paper.
\item Zugner and Gunnemann~\cite{zugner2019adversarial}: $O(N^2)$, as shown in the paper.
\end{itemize}

\end{document}